\documentclass{article}

\usepackage[acronym]{glossaries}
\usepackage{microtype}
\usepackage{graphicx}
\usepackage{subfigure}
\usepackage{booktabs} %

\usepackage{hyperref}
\usepackage{tikz}
\usetikzlibrary{positioning}
\usetikzlibrary{calc}

\usepackage[accepted]{icml_arxiv}

\usepackage{amsmath}
\usepackage{amssymb}
\usepackage{mathtools}
\usepackage{amsthm}
\usepackage{enumitem}   

\usepackage[capitalize,noabbrev]{cleveref}

\theoremstyle{plain}

\theoremstyle{definition}

\theoremstyle{remark}

\newacronym{vla}{VLA}{Visual Language Action}
\newacronym{fvwm}{FVWM}{Foundation Veridical World Model}
\newacronym{fwm}{FWM}{Foundation World Model}

\usepackage[textsize=tiny]{todonotes}

    \icmltitlerunning{The Essential Role of Causality in Foundation World Models for Embodied AI}

\begin{document}

\twocolumn[
\icmltitle{
The Essential Role of Causality in Foundation World Models for Embodied AI
}

\icmlsetsymbol{equal}{*}

\begin{icmlauthorlist}
\icmlauthor{Tarun Gupta}{equal,ms}
\icmlauthor{Wenbo Gong}{equal,ms}
\icmlauthor{Chao Ma}{equal,ms}
\icmlauthor{Nick Pawlowski}{equal,ms}
\icmlauthor{Agrin Hilmkil}{equal,ms}
\icmlauthor{Meyer Scetbon}{equal,ms}
\icmlauthor{Marc Rigter}{equal,ms}
\icmlauthor{Ade Famoti}{ms}
\icmlauthor{Ashley Juan Llorens}{ms}
\icmlauthor{Jianfeng Gao}{ms}
\icmlauthor{Stefan Bauer}{tum}
\icmlauthor{Danica Kragic}{kth}
\icmlauthor{Bernhard Schölkopf}{mpi}
\icmlauthor{Cheng Zhang}{equal,ms}
\end{icmlauthorlist}

\icmlaffiliation{ms}{Microsoft Research}
\icmlaffiliation{tum}{Technical University of Munich}
\icmlaffiliation{mpi}{ Max Planck Institute for Intelligent Systems}
\icmlaffiliation{kth}{KTH royal institute of technology}

\icmlcorrespondingauthor{MSR Causica Team}{causica@microsoft.com}

\icmlkeywords{Machine Learning, ICML}

\vskip 0.3in

]

\printAffiliationsAndNotice{\icmlEqualContribution} %

\begin{abstract}
Recent advances in foundation models, especially in large multi-modal models and conversational agents, have ignited interest in the potential of generally capable embodied agents. Such agents will require the ability to perform new tasks in many different real-world environments. 
However, current foundation models fail to accurately model physical interactions and are therefore insufficient for Embodied AI.
The study of causality lends itself to the construction of veridical world models, which are crucial for accurately predicting the outcomes of possible interactions. This paper focuses on the prospects of building foundation world models for the upcoming generation of embodied agents and presents a novel viewpoint on the significance of causality within these. We posit that integrating causal considerations is vital to facilitating meaningful physical interactions with the world. Finally, we demystify misconceptions about causality in this context and present our outlook for future research. 
\end{abstract}

\section{Introduction}

The ability to perform physically meaningful interactions in real-world environments is essential to maximizing the potential of technology in everyday life. Entities capable of conducting interactions of this kind are in our work referred to as \textit{embodied agents}. These agents can vary in form, ranging from mechanical robots to humans augmented by AI in mixed or virtual reality environments where interactions bear physical importance.

    Foundation models account for a significant advancement in generalization across a range of tasks with multi-modal input, compared to canonical deep learning models from the past decade. A single foundation model 
can address a wide variety of tasks in a zero-shot manner.
For example, in robotics, we anticipate that foundation models for embodied agents will make the introduction of new robots into our life effortless: each new robot can immediately solve a given task using the same model without the need for a specialized model.
In addition, robots can quickly adapt to a new environment by interacting with it and using the new experiences to improve without lengthy retraining or complex programming.

However, contemporary foundation models are not sufficient.
Current approaches, dominated by large (vision-) language models \cite{achiam2023gpt,bubeck2023sparks,chen2023llava}, are based on correlational statistics and do not explicitly capture the underlying dynamics, compositional structure or causal hierachies. The lack of a veridical world model renders them unsuitable for use in Embodied AI, which demands precise or longterm action planning, efficient and safe exploration of new environments or quick adaptation to feedback and the actions of other agents. 

World models~\cite{ha2018world} make action-conditioned predictions about the future state of the environment by training on action-labelled interaction data. World models therefore capture the consequences of actions and are suitable for optimising decision-making~\cite{sutton1991dyna, janner2019trust}. To train general world models it will be necessary to leverage large readily-available observational datasets, such as internet videos, in addition to action-labelled data which is much more scarce and expensive to generate. The best way to utilise observational data for world modeling remains an open question~\cite{seo2022reinforcement, bruce2024genie}. In this paper we argue that causal considerations are necessary to unlock the potential for training world models from internet-scale data.

Causality at its core aims to understand the consequences of actions, allowing for interaction planning. Philosophy and cognitive sciences sees understanding causal concepts as both the basics and sometimes the ultimate goal for humans to learn how to interact with the world \citep{Gibson1978ecological, gopnik2007causal, adams2010causal} %
and key in children's development \citep{piaget1965stages,gibson1988exploratory}. This aligns with the concept of affordances in Embodied AI research \cite{kjellstrom2011visual,koppula2013learning, koppula2013anticipating, ardon2020affordances, ahn2022can}, which pertains to the types of action that can be applied and the resulting consequences.
Importantly, even with the help of available real or simulated environments, the experimentation might still be too coarse-grained to deal with spurious relationships \citep{herd2019detecting, lavin2021simulation}.
Thus, the world model for an agent, including humans, should be causal-aware \citep{anonymous2024robust} and able to efficiently incorporate data from observations, demonstrations, and interaction (i.e., the property referred as \emph{veridicality} in this paper). 
As such, causality offers tools and insights that hold the key pieces to building \textbf{\glspl{fvwm}} (Section \ref{sec:FVWM}) that will power future embodied agents.

In this position paper, we argue that future progress in Embodied AI relies on further advances in foundation models (Section \ref{sec:eai}). For such research, causality is pivotal in developing the \gls{fvwm}, which requires us to go beyond canonical causal research paradigms (Section \ref{sec:causal}).  Thus, we foresee research opportunities (Section \ref{sec:research}) advancing foundation world models in Embodied AI from a causal perspective. 
We extend our discussion with considerations for the eventual deployment of Embodied AI (Section \ref{sec:impact}).

\section{Embodied AI and Foundation Veridical World Models}
\label{sec:eai}

\begin{figure}
    \centering
    \resizebox{\linewidth}{!}{
\begin{tikzpicture}[
    block/.style={
        draw,
        fill=blue!20,
        rectangle,
        rounded corners,
        text width=2cm,
        align=center,
        minimum height=2cm
    },
    title/.style={
        draw,
        fill=blue!20,
        rectangle,
        rounded corners,
        text width=10cm,
        align=center,
        minimum height=1cm
    }
]

\node (title) [title]  {Foundation Veridical World Model};

\node (app1) [block, below=of title, yshift=0.5cm, xshift=-3.9 cm] {Daily assistance humanoid};
\node (app2) [block ,below=of title, right=of app1, xshift=-0.7cm] {Industrial manufacturing robots};
\node (app3) [block, below=of title, right=of app2, xshift=-0.7cm] {Mixed-reality interactions};
\node (app4) [block, below=of title, right=of app3, xshift=-0.7cm] {... ...};

\end{tikzpicture}
}
\vspace{-5mm}
    \caption{The Foundation Veridical World Model (FVWM) is designed to comprehend associations, counterfactuals, and interactions within the world, aiding any embodied AI agent in executing tasks across diverse environments and platforms. %
    }
    \label{fig:FWM}
\vspace{-5mm}
\end{figure}
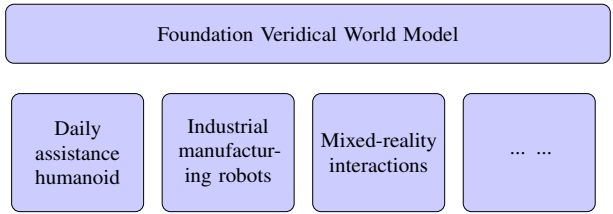

For many years, research has pursued the creation of Embodied AI. 
However, existing approaches are  limited to performing narrow tasks in highly constrained environments, such as assembling machines in production lines. This inhibits the applicability of existing Embodied AI systems to general real-world settings, where environments are often dynamic and unstructured, and the set of tasks to be performed is open-ended.
This highlights a significant gap between current technologies and  fulfilling the promises of general Embodied AI.
We believe that recent technological advancements (Section \ref{sec:readyEAI}), including hardware, foundation models, and data, have brought general Embodied AI within reach, where the remaining gap can be overcome by veridical world models (Section \ref{sec:FVWM}).

\subsection{The readiness of Embodied AI research}
\label{sec:readyEAI}

 \paragraph{Hardware advancement} The pace of innovation in hardware and systems, ranging from real-world robots to virtual and mixed reality devices, is advancing at an unprecedented rate, though there is still a significant journey ahead before achieving maturity for general tasks. 
The costs of both humanoid (e.g.~Optimus, Phoenix~\cite{Phoenix}, Digit~\cite{digit}) and quadrupedal robots (e.g. Spot~\cite{Spot}, Go2~\cite{Go2} ) have been rapidly decreasing over the last decade, with a huge potential market size \cite{GoldmanSachs,tilley2017automation}. For virtual mixed reality experiences, Oculus from Meta has increased in popularity and Apple's Vision Pro is taking a commercial leap into offering new interactive experiences, creating opportunities for virtual agents. In addition, new types of sensors for robots have been invented recently. Electronic skin can capture the sensation of touch \cite{stern2021electronic} to enable dexterous robots~\cite{lambeta2020digit, sun2022soft}, while perovskite retinomorphic sensors \cite{trujillo2020perovskite} mimic the human retina and react to changes in illumination rather than constant signals. These advancements offer different ways to sense the world. 
This enables the consideration of more dextrous manipulation tasks, such as those involving deformable objects, rather than rigid ones \cite{yin2021modeling}.

Meanwhile, the evolution of AI computing power has accelerated at an unprecedented rate. This surge in computing capabilities is primarily driven by innovations in hardware, such as more powerful GPUs and specialized AI processors \cite{dally2021evolution, saravanan2023advancements}, providing the computing basis for foundation models. These developments have enabled AI systems to process vast amounts of data more quickly and more effectively than ever before, expanding the horizons of potential across various domains.

\paragraph{Paradigm shift in AI} The paradigm shift driven by foundation models in both language and vision has markedly enhanced the ability of AI models to understand and perceive the environment \cite{bubeck2023sparks,achiam2023gpt,yuan2021florence,liu2023visual,chen2023llava}. These systems now demonstrate an increasing capability to generalize on high-level planning and decision-making tasks, often in a zero-shot manner~\cite{ahn2022can}. This has led to a shift in the research paradigm from specialized models for a single target task and dataset to foundation models generalizing across many tasks and datasets. These increased generalization capabilities hold the key to building  world models suitable for general Embodied AI systems. 

\paragraph{Data availability} Adding to the timeliness of this venture is the increasing amount of of data in the field of Embodied AI, which is crucial for effective training of foundation world models. 
Robotics simulation environments and demonstration datasets have become more diverse, encompassing different types of tasks and scenarios, such as OpenX embodiment \cite{padalkar2023open}, Isaac gym \cite{makoviychuk2021isaac,ma2023eureka}, CloudGripper \cite{zahid2023cloudgripper}, Habitat 3.0 \cite{puig2023habitat} and Causal World simulations \cite{ahmed2020causalworld,rrc2021}, greatly increasing the amount of available data. In addition, the maturity of physical robots and sensors also facilitates the generation of diverse multi-modal data. These real-world datasets represent different aspects of the environment. Moreover, sources like YouTube provide a wealth of information in the form of videos or general human behaviors~\cite{seo2022reinforcement}.

\subsection{Towards Foundation Veridical World Models}
\label{sec:FVWM}
\begin{figure}[t!]
    \centering
    \centering
    \resizebox{\linewidth}{!}{
\input{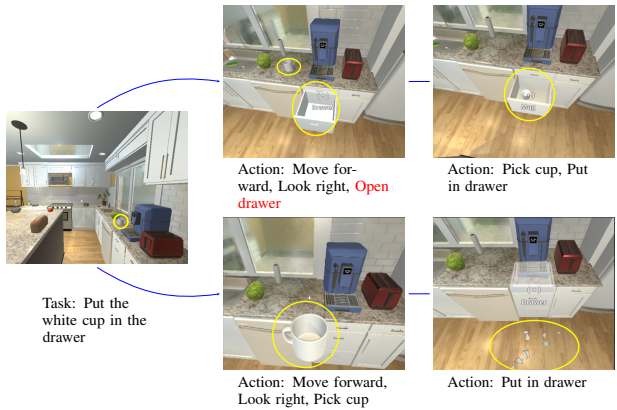}
}
\vspace{-5mm}
    \caption{A veridical world model can predict the consequences of a sequence of actions and corresponding counterfactuals correctly with confidence, ignoring irrelevant information like color of the cup. For example, the model should correctly predict that the cup will break if `open drawer' action is not taken in the sequence. Images taken from AI2THOR \cite{kolve2017ai2} simulator. }
    \label{fig:enter-label}
\vspace{-5mm}
\end{figure}

We believe that world models hold great promise for the future of Embodied AI. Ideally, a general world model will facilitate the entire Embodied AI pipeline, spanning various tasks and platforms as shown in Figure \ref{fig:FWM}. To achieve that, a required property is the ability to consume multi-modal inputs, and generalize across environments, domains, and tasks (i.e., being \emph{foundational}).  This requires the model to faithfully understand and model the world dynamics, i.e., being \emph{veridical}. It needs to go beyond merely summarizing the content given what is \emph{observed}, but also truly comprehend the world in a way that enables actionable \emph{interventions}.

Combining the above two key aspects leads to the idea of \textbf{\textit{\glspl{fvwm}}}, or simply \textbf{\textit{\glspl{fwm}}}, a single (or set of) veridical, multi-modal models that can: 
\begin{enumerate}[label=(\roman*)]
\setlength\itemsep{0pt}
    \item (Representation) Conceptually understand the components, structures, and interaction dynamics within a given system at different levels of abstractions \cite{olsen2023social};
    \item (Veridicality) Quantitatively model the underlying laws of such a system, that enables accurate predictions of counterfactual consequences of interventions/actions;
    \item (Foundational) Generalize (i) and (ii) across diverse systems or domains encountered in the world.
\end{enumerate}

When only (i) and (ii) are satisfied, \glspl{fvwm} reduces to world models, originally intended to simulate possible futures of a single system \cite{ha2018world, li2020causal}. To accomplish (iii), \glspl{fvwm} require a flexible formulation that can, using observational and interaction data from diverse real-world datasets, reveal the underlying principles that generalize across environments. This naturally falls into the realm of causality. However, incorporating such causal reasoning abilities while being able to process and generalize across multi-modal datasets and tasks is challenging, indicating substantial research opportunities for significant real-world impact.

\section{Causality in Foundation World Models}
\label{sec:causal}

In this section, we will demystify the next direction in causal research to be used for \glspl{fvwm}. Specifically, we will: (i) briefly review SEMs, one of the commonly used canonical causal approaches; (ii) discuss the limitations of these canonical approaches in the context of \glspl{fvwm};  (iii) describe the necessities of identifying new paradigms of causality research for \glspl{fvwm} by demystifying a few common misconceptions.

\subsection{Canonical approaches to causal modelling}
\label{sec:currentcausal}
Historically, causal research has been compartmentalized into distinct tasks. Causal discovery, for example, aims to recover causal relationships among variables from a dataset \cite{glymour2019review,spirtes2000causation}, while causal inference is concerned with quantifying the effects of interventions. The advent of causal machine learning has expanded these horizons, introducing approaches such as end-to-end causal inference \cite{geffner2022deep}, alongside advancements in causal representation learning \cite{scholkopf2021toward} and causal reinforcement learning \cite{zeng2023survey}. There are two main frameworks to describe causality: potential outcomes (PO \cite{rubin2005causal, hernan2010causal, imbens2015causal}) and Structural Equation Models (SEMs) \cite{pearl2009causality, peters2017elements}. In the following, we briefly introduce the concept of SEM.

\paragraph{Structural Equation Models (SEM)}  
SEMs describe the generating mechanisms between variables.  
This typically follows the form of $$X_i = f_{X_i}(\text{Pa}_{X_i},U_{X_i}),$$
where $f_{X_i}$ is a function mapping the values of exogenous variables $\mathcal{U}_{X_i}$ and its causes $\text{Pa}_{X_i}$ to the observation $X_i$.

A fully specified SEM generates a range of distributions that are systematically organized within the Pearl Causal Hierarchy (PCH) \cite{pearl2000models}, which categorizes causal information into three distinct levels: observational, interventional, and counterfactual. These levels reflect an ascending order of causal depth, each enabling a richer understanding of causality.  
\begin{enumerate}[label=(\roman*)]
\setlength\itemsep{0pt}
    \item The \textbf{observational} level pertains to the distribution of observation variables without any interventions, typically represented as $p(\mathcal{X})$. It provides insights into natural associations and correlations.%
      
    \item The \textbf{interventional} level involves altering the system through specific actions, indicated by Pearl's $do$ operator.  
    The interventional distribution, $p(\mathcal{X}_{-i} \, | \, do(X_i = a))$, characterizes the consequence of the action.   
      
    \item The \textbf{counterfactual} level delves into hypothetical scenarios, considering what would occur had a different action been taken, given the same initial circumstances, 
    expressed as $p(X_j \, | \, do(X_i = a), \mathcal{X}_{obs}, \mathcal{U}_{obs})$. 
\end{enumerate}
  
The PCH provides a framework for reasoning in a causal-aware manner, allowing a model not only to observe the world but to interact with it as well.

\paragraph{Theory-oriented research paradigm}  
Given the difficulty of accessing the true system, theory-oriented research prevails in causality. %
For example, identifiability theory concerns the theoretical feasibility of deducing causal relations, interventional or counterfactual effects from observational data, assuming a certain set of models. %
Once identifiability is determined, the next steps involve developing methods for identifying \cite{pearl2009causal, angrist1996identification} and accurately estimating these causal quantities \cite{rosenbaum1983central, stuart2010matching, li2018balancing, chernozhukov2018double, wager2018estimation}.
This theory-driven approach is one way to ensure the precision and trustworthiness of the model, as it contributes significantly to our comprehension of underlying causal mechanisms and guides decision making in relevant domains. However, as we shall discuss, this approach may limit the real-world impact of methods developed~\cite{imbens2020potential}.

\paragraph{Limitations}  
The canonical theory-oriented research has facilitated significant progress. Meanwhile, we are at a historical juncture to rethink this strategy for \glspl{fvwm} and embodied AI due to the following limitations. %
First, real-world sensory inputs are usually not structured into well-defined variables that admit a causal explanation (in terms of SEM/PO) and meaningful interventions \cite{scholkopf2021toward}. Second, in many real-world scenarios, certain key theoretical properties, such as identifiability, may not hold, and their associated assumptions are often not verifiable \cite{pearl2009causality, robins2000marginal}.   
Third, in the context of \glspl{fvwm}, the case-specific nature of causal identification complicates the integration of causal reasoning \cite{bareinboim2022pearl}. Foundation models aim to provide a generalizable understanding of the world, but the bespoke nature of causal identification, which often relies on domain-specific assumptions, poses a significant barrier for general environments. This discordance reveals a gap between the theoretical guarantee of causal reasoning and the practical needs of causal principles in foundation models, necessitating a re-evaluation of the approaches used to embed causality for \glspl{fvwm} and embodied AI.

\subsection{Demystifying causal modeling for \glspl{fvwm}}
\label{sec:myth}

The canonical causal research paradigm has inadvertently contributed to certain misconceptions regarding the field of causal machine learning, thus limiting its broader impact, especially in the context of \glspl{fvwm} and embodied AI. In the following, we aim to demystify three main misconceptions: (1) casual research is a separate field from Embodied AI and \glspl{fvwm}; (2) causal research must be theory-oriented; and (3) predictive and causal models are different.

\paragraph{Misconception I: Causal ML is a separate field from Embodied AI and \glspl{fvwm}.}
Despite causality being pivotal for the generalization of world models \cite{anonymous2024robust}, Causal ML is often overlooked in Embodied AI discussions, compared to RL and foundation models. This oversight partly stems from distinct tools, such as SEMs and PO, and the application domains traditionally associated with canonical causal research. %
However, it is crucial to recognize that Causal ML is intricately connected to both reinforcement learning (bandits and imitation learning as a special case) and foundation models, with shared goals. 

Bandits and online RL share the aim of learning the best action policy from online interactions. Offline RL assumes access to only fixed dataset from the system of interest, and therefore considers the same problem setting as causal modeling~\cite{kaddour2022causal}. 
While RL does not always explicitly model environmental dynamics -- as seen in model-free RL -- it implicitly captures these dynamics within the learned policy or value function. 
Thus, RL is intrinsically causal in the sense that it needs to inherently understand the consequences of actions for policy learning.
A more explicit connection can be seen in model-based RL, where temporal causal models can be used to formulate the transition dynamics. 
Such temporal causal models induce the transition probabilities between states across time. 
Actions by agents are then viewed as interventions to this model. 
Leveraging ideas from causality would allow the learned dynamics to only reason over causally relevant attributes such as force and location to predict the next state (e.g.~location of the cup in \cref{fig:enter-label}). 
Many recent works have begun to incorporate causal concepts into the state transition dynamics (see \cref{sec:related}).
This potentially enhances sample efficiency and generalizability beyond training contexts.

Large multi-modal 
foundation models aim to facilitate reasoning and planning \cite{sun2023survey, huang2022towards, webb2023emergent, li2023multimodal}, which inherently requires understanding the physical nature of environments. Although much progress has been made \cite{wei2022chain, zhang2022automatic,creswell2022selection,wang2023plan,green2012art}, they are trained to gather information only at the observational level, which may limit their understanding of the physical world (lack of interventional and counterfactual information). While these models, including CLIP \cite{radford2021learning} and GPT-4 \cite{achiam2023gpt}, excel in various tasks like vision-language understanding and image recognition with multi-modal input, 
they are limited by issues like hallucination \cite{huang2023survey, gunjal2023detecting} and reduced multi-step reasoning abilities \cite{jin2024impact, sun2023survey}, indicating the lack of causal understanding.
On the other hand, canonical causality research fails to incorporate multi-modal datasets and is typically tied with predefined variables for SEM. 
Recent work (discussed in \cref{sec:related}) reveals the synergy between Causal ML and foundation models for mutual enhancement, and paves the way for \glspl{fvwm}.

\paragraph{Misconception II: Causal ML research must be theory-oriented.}

The traditional emphasis on %
theory-oriented approaches in causal machine learning research  
often overshadows empirically-driven approaches, which prioritize causal experimentation, verification, and hypothesis selection
\cite{dorie2019automated, jensen2019comment, gong2022instructions, nie2021quasi, saito2020counterfactual, vansteelandt2012model,schuler2018comparison, rolling2014model, matthieu2023select,cook2002experimental,hoover2001causality,bareinboim2016causal, bareinboim2014generalizability}. 
Consequently, over-reliance on purely theoretical approaches can lead to models that fail to perform effectively in practical, complex scenarios. We would like to emphasize that one of the end goals of causality research is causal-aware reasoning and the ability to predict the consequences of any action accurately. Canonical formulations with theoretical guarantees are one approach to causal research, while focusing on empirical evaluations of causal properties is an alternative when it comes to \glspl{fvwm}. 
A shift towards empirically-driven research allows models to assimilate causal knowledge from a wide array of scenarios, akin to the way large language models are trained and evaluated.
Moreover, the unique problem setting of Embodied AI opens up the possibility for empirical evaluation to become the cornerstone of causal model assessment. Through interaction with both real and simulated environments, Embodied AI agents can generate and leverage interventional or counterfactual data, thereby permitting a more pragmatic approach to causal understanding. Thus, empirically-driven causal research under \glspl{fvwm} and embodied AI is equally or more important than the theory-oriented paradigm.

\paragraph{Misconception III: Predictive models cannot be causal.}
The principle that correlation does not imply causation is highly relevant in the field of predictive modeling. It is often misunderstood as meaning that predictive models cannot be causal. This interpretation holds some truth when models are trained solely on observational data, which are prone to confounding effects that can bias predictions about the outcomes of specific actions. 
However, it does not apply when a predictive model is trained with  active intervention data, which allows the model to demonstrate causal properties. This aligns with the discussion in misconception I : RL models are inherently causal and can learn to take optimal actions for a variety of tasks \cite{kaddour2022causal,anonymous2024robust}. 
Thus, causal models are inherently high-accuracy prediction models of consequences about all possible actions. Predictive models can therefore be considered causal models when trained with appropriate algorithms and datasets, even without explicit causal theory considerations. In the realm of Embodied AI, opportunities arise for gathering suitable interaction data in the physical world. This enables the integration of both causal considerations and large-scale prediction models \cite{ke2019learning, scherrer2021learning, toth2022active}. 

\section{Research opportunities}
\label{sec:research}
Recognizing the immense potential of foundation models for Embodied AI and the necessity of a causal perspective to ensure accurate estimation of action consequences, we foresee significant research opportunities, especially since current foundation models are not causal-aware. In particular, we identify a pressing need for research advancement in the following aspects.

\paragraph{Diverse modalities}
The foundation model for Embodied AI extends beyond traditional vision and language realms. Robotics necessitates the integration and interpretation of diverse sensory data, including tactile and torque sensors, covering a broad range of modalities which vary from one agent to the next. This requirement surpasses the capabilities of current vision and language-based foundation models. Developing a system capable of effectively processing and utilizing these complex sensory data is crucial for enhancing robotic capabilities, especially in more intricate and nuanced tasks. On the other hand, building a causally-aware \glspl{fwm} that comprehends high-dimensional observations from a large number of potential different modalities to model the underlying mechanisms is very challenging and presents multiple research opportunities.

First, integrating data from multiple modalities to achieve a cohesive understanding from different sensory inputs poses challenges in spatial and temporal alignment \cite{ward2018tactip,lin2023bi,wang2023skyscript,huang2023voxposer,behley2019semantickitti} due to the varying perspectives and information overlap provided by each sensor, as well as the need to reconcile different sampling intervals to maintain an up-to-date representation of the world.

Second, agents vary widely in their sensor and effector configurations, from humanoid robots like Tesla's Optimus to quadrupedal robots like Boston Dynamics' Spot, to virtual cars, each with unique sensory inputs and available effectors. For \glspl{fwm} to make precise action-outcome predictions, they must semantically interpret these diverse inputs and controls, necessitating a representational model that encapsulates the essence of an agent's design to enable generalization across different configurations.

Lastly, handling the complexities of varying abstraction levels for effective interaction presents a formidable challenge \cite{scholkopf2021toward,wang2021desiderata,koul2023pclast,zhang2020causal,lamb2022guaranteed}. In standard causal models, variables are precisely defined within structural equation models (SEMs) for a single dataset and task. However, such clarity is often absent in the realm of multi-modal, high-dimensional inputs across diverse tasks. While emerging works are exploring causal-aware foundation models \cite{zhang2023towards, ke2022learning}, they remain confined to tabular data with predefined tasks.
Given a high-dimensional input, a representation meaningful for interactions and their effects (affordances) is necessary to ensure the world model is meaningful. For instance, modeling interaction and consequences at a pixel level provides little immediate insight and requires additional transformations to either map the (motor) control of an agent into actions in pixel space or to extract an interpretable understanding of the effect the interaction had. Additionally, a foundational world model for embodied AI, which aims to achieve broad generalization, must consider different knowledge hierarchies to ensure effective generalization and knowledge transfer across varied contexts. 
For example, understanding the consequences of avoiding a car attempting to overtake should be consistent regardless of the specific car instances (car, brand), and the same principle applies to home robotics with different effectors for similar tasks. The FVWM, therefore, requires hierarchical representations that facilitate knowledge transfer at multiple levels, enabling nuanced understanding and application across various scenarios and tasks.   

\vspace{-2mm}
\paragraph{New paradigm for interactions}
As discussed, canonical research in causality and RL shares similar objectives but differs in the source of the data: observational data versus interventional. Traditional causality research seeks to learn the world model using existing data, assuming that new interventions are not feasible, while online RL learns through interactions. For Embodied AI with FVWM, there will be access to a vast amount of offline observational data from humans, such as vision language data, human motion data, and robotics data, along with the possibility to obtain new intervention data. We need to utilize all different data paradigms, breaking down the barriers between foundation models, RL, and causality.

Pre-training FVWMs from offline data alone may be sufficient to obtain good performance in some scenarios. However, we anticipate that gathering additional online interaction data~\cite{lee2022offline} will be necessary to obtain robust performance across general tasks and environments. This data should be collected in a way such that it enables the model to reason about interventional and counterfactual scenarios and therefore integrating different levels of causal reasoning \citep{pearl2009causality}. Therefore, a key research direction will be developing algorithms that efficiently gather the most useful intervention data online, enabling fast adaptation:
\begin{itemize}
    \item The related areas of active learning and experimental design selectively query datapoints to maximize information gain \citep{gonzalez2016glasses,ma2019eddi,gong2019icebreaker, yin2020reinforcement}. Similar ideas have been explored in the context of training world models by collecting data where the world model has the greatest epistemic uncertainty \citep{sekar2020planning, rigter2024reward}. However, maximising uncertainty does not necessarily lead to the collection of data that is most useful for improving decision-making \citep{pathak2017curiosity}. Another paradigm then is to incentivize agents to collect semantically diverse data from online interactions based on representations grounded in natural language \citep{gupta2022foundation, tam2022semantic} or control-relevant information contained in observations \citep{lamb2022guaranteed, islam2022agent}. 
    \item The research into unsupervised environment design \citep{dennis2020emergent, samvelyan2023maestro} aims to learn to adapt the difficulty of the environment to the agent's learning progress and can produce a natural curriculum of increasingly complex environments for agents to interact with and collect data from, thereby improving the knowledge of world model in a continual manner. 
\end{itemize}
In summary, by training a causally aware model, FVWMs may unlock opportunities to develop algorithms that efficiently gather the intervention data that is most relevant for improving decision-making. Such an approach has the potential to greatly expand the generality and reliability of the world model while reducing the cost of online data acquisition.

\paragraph{Improving planning} FVWMs aim to facilitate long-term decision-making in Embodied AI systems. It is well-known that improved prediction accuracy leads to improved planning \citep{kearns2002near, janner2019trust}. However, even with an accurate world model, optimising behaviour with large prediction models is still a challenging problem and requires either reinforcement learning \citep{hafner2020mastering}, search \citep{schrittwieser2020mastering}, or sophisticated model-predictive control \citep{hansen2022temporal} within the world model. While these approaches have proven successful in narrow tasks, it is unclear whether they will scale to foundation world models trained across diverse tasks and environments. By modelling causal structure, FVWMs have the potential to learn representations that are highly suitable for planning and optimisation. Thus, investigating how the representations learnt by FVWMs can improve not only prediction accuracy and generalization, but also planning, is another key research direction.

\paragraph{Latent Dynamic Representations for World Models}
To enable sample efficient learning and planning with strong generalization abilities and causal reasoning, it will be crucial for FVWMs to learn suitable abstract representations of the world. Current large world models such as GAIA \citep{hu2023gaia}, LWM \citep{liu2024world} learn to encode the entire images into a latent space based on a reconstruction loss which can capture many irrelevant details for decision-making. Therefore, minimal state representations \citep{lamb2022guaranteed, islam2022agent}, which only capture agent controllable features of the environment, are more suitable for causal reasoning and planning. Some possible approaches are using regularisation techniques such as sparsity or causal priors \citep{li2024enforcing, killedar2021learning} in the latent space, which can be thought of as encouraging the latent space to learn the SEM \citep{rajendran2024learning}. Other possibilities include changing the model architecture to learn latent representations that disentangle task-irrelevant visual details from task-relevant details and dynamically adapt these representations as task distribution of interest changes. With improved state representations, we expect that the model should be able to capture causal relationships and therefore learn in a more sample-efficient way and generalise better.

\vspace{-2mm}
\paragraph{Empirically-driven research and evaluation}
The goal of Embodied AI research is to improve the performance and applicability of embodied agents in real-world scenarios. Therefore, to measure success we should focus on empirical performance on metrics related to physically meaningful tasks, rather than theoretical results. Thus, shifting the research culture towards being empirically-driven, especially for causal researchers, and developing relatable, realistic, and economical evaluation methods is crucial for success.

Evaluating real-world scenarios with a physical system is very costly in terms of time and potential risks, although it is necessary for the ultimate success of Embodied AI. Thus, constructing a curriculum of tasks that are easy to evaluate is key. This means evaluations can be performed at scale, replicated by any research institute, have clear metrics demonstrating progress, and provide a reliable approximation for end tasks in the real world. This requires not only further advancements in Embodied AI in simulated environments like CausalWorld \cite{ahmed2020causalworld, rrc2021} and Habitat 3.0 \cite{puig2023habitat} but also the creation of innovative tasks that aid in this process. For example, evaluation on counterfactual prediction can be a good approximation for assessing true world understanding. However, we need to consider concrete tasks and evaluation metrics. In the case of counterfactual prediction, the task could involve generating a video \cite{yin2023nuwa}, while the metric needs to be nuanced as pixel-space evaluation has drawbacks~\cite{fernandez2018new}.%

Moreover, it's unlikely that a single metric will fully capture the model's performance, especially with the new causal perspective. Traditional metrics such as graph prediction accuracy or treatment effect estimation accuracy with well-defined variables become less relevant. However, the philosophical approach and perspective in evaluating causal properties remain crucial. For example, it's worth considering if evaluating causal metrics in a representation space will provide additional insights complementary to prediction tasks. Also, the key to Embodied AI is understanding affordances and the consequence of actions, so evaluating counterfactuals in an imagined world might be effective.

On the other hand, the primary reason that correlation-based models are often considered unsuitable for causal-aware tasks is the bias present in the datasets collected. A causal-aware model is expected to accurately predict future outcomes in out-of-distribution evaluations (similar to A/B testing data), even when trained with biased data. Therefore, it's crucial to consider the availability of training data and the types of properties that reflect real-world scenarios, while also ensuring that the evaluation dataset genuinely represents generalization across various scenarios.

In conclusion, while the broader concept of Embodied AI encompasses virtual and mixed reality environments, as well as physical real-world environments (robotics and human action) , the final evaluation, especially considering human safety, cannot be overlooked \cite{kadian2020sim2real,ma2023high,Jinetal23,yang2023pave}. All these efficient evaluation methods aim to pave the way for real-world deployment.

\section{Deploying Embodied AI}
\label{sec:impact}

Many factors affect how \glspl{fvwm} can impact the deployment of Embodied AI in the real world. In this chapter, we focus on what we believe will be one of the most important factors, which is \glspl{fvwm} ability to greatly scale up the deployments of Embodied AI in robotics.

\vspace{-2mm}
\paragraph{General purpose robots}
General purpose robots form perhaps the most exciting prospect of generally capable Embodied AI. By compatibility with human environments and tools, robots of this kind could finally bridge the gap between the physical and digital worlds and allow large-scale automation of physical tasks. Potential applications include  automating domestic tasks~\cite{puig2023habitat}, nursing~\cite{ohneberg2023assistive}, work in hazardous environments~\cite{petereit2019robdekon}, and manufacturing~\cite{kheddar2019humanoid}. We discuss ethical considerations regarding societal impact in our impact statement. Recent work has demonstrated zero-shot solutions to complex tasks, yet shows poor performance when faced with new environments. Research in causal \glspl{fvwm} should prioritize consideration of diverse modalities and platforms, enabling success across a wide range of tasks and environments.

\vspace{-2mm}
\paragraph{Rapid deployment of specialized robots}
While general purpose robots capable of zero-shot learning stand to reduce deployment time, specialized robots will remain more cost effective in some scenarios - especially for large-volume production.
Examples include farming~\cite{aravind2017task}, scientific lab exploration~\cite{ren2023autonomous,boiko2023autonomous} and industrial welding machines~\cite{grau2020robots}. Implementing specialized robots is a time-consuming process and therefore learning to perform tasks without explicit programming an important research area~\cite{dzedzickis2022industrialrobotics}. However, current learning implementations are typically limited to a single environment~\cite{meyes2017motion}. We believe that causal-aware \glspl{fvwm} may resolve these constraints and therefore reduce the cost to deploy specialized industrial  robots. An important testbed for this work will therefore be operation across robot morphologies with different sensing and tooling attachments.

\vspace{-2mm}
\paragraph{Robustness and safety}
Industrial robots have typically been confined to isolated cells. However, constraints such as force limitations have enabled more collaborative operation, improving productivity~\cite{dzedzickis2022industrialrobotics}. In other applications, such as nursing homes, direct physical interaction with humans is vital. Yet in such situations, simple threshold constraints may not be sufficient to ensure physical safety while still allowing the robot to perform the tasks of interest. A thorough grasp of the outcomes resulting from physical interactions, as facilitated by \glspl{fvwm}, presents an opportunity to enhance safety by developing more nuanced safety constraints. 

\section{Related Work}
\label{sec:related}
\paragraph{Foundation model for Embodied AI}
There have been a number of surveys on foundation models with a focus on recent research advances, challenges, and opportunities \citep{zhou2023comprehensive, bommasani2021opportunities}. Regarding Embodied AI, \citet{firoozi2023foundation, xiao2023robot} survey the role of pretrained foundation models in robotics for perception, navigation, decision making and control. Recent studies have explored the use of vision-language FMs in Embodied AI tasks to improve sample efficiency and performance abilities \citep{khandelwal2022simple, batra2020objectnav, shridhar2022cliport}. Moreover, numerous surveys from the perspective of robotic tasks have been published, providing insights applicable to both deep learning and machine learning at large \cite{yin2021modeling, newbury2023deep}. Recent research advances have increasingly focused on utilizing foundation models for Embodied AI. A number of methods leverage large LLMs and VLMs for planning, re-planning and navigation  \cite{lin2023text2motion,schumann2023velma,skreta2024replan}. 
Also, more recent work focuses on generalization across different environments and tasks for robots based on VLMs, such as SayCan \citep{brohan2023can}, RT-2 \citep{brohan2023rt}, GATO \citep{reed2022generalist}, AutoRT \citep{ahn2024autort}, and autonomous driving, such as GAIA \citep{hu2023gaia}.

\vspace{-3mm}
\paragraph{Causality and Causal Foundation Model}
In recent years, books \cite{pearl2009causality,hernan2010causal,cunningham2021causal, imbens2015causal, spirtes2000causation,peters2017elements} as well as surveys \cite{spirtes2016causal, kitson2021survey, shojaie2022granger, zanga2022survey} have been published, drawing attention to causality and causal machine learning in general.  Related to our work is the recent overview of open problems from 
\cite{kaddour2022causal}, which connect different machine learning approaches to causality \citep{Scholkopf22b}; and a survey and position paper on causal reinforcement learning \cite{zeng2023survey} which aligns with our position of breaking the boundary of between these fields. There is large body of research work considering generic deep learning and causality including \cite{scholkopf2021toward,jin2022causalnlp, zhou2023opportunity,berrevoets2023causal}.

With the advent of foundation models, particularly large language models (LLMs), there have been multiple position and evaluation papers considering whether LLMs are causally aware and if they can be used to complete causal tasks \cite{willig2022can, tu2023causal, kiciman2023causal, zhang2023understanding, zhang2023causality, jin2023can, Jinetal23, zevcevic2023causal}. Generally, these papers argue that existing LLMs can discuss causality but are not capable of reasoning in a causally aware manner. Consequently, this further confirms that current LLMs are not yet world models. Meanwhile, recent research has begun to show that it may be possible to innovate based on transformer-style architecture \cite{lorch2022amortized, zhang2023towards}, enabling foundation models to estimate treatment effects. Although these works are in their early stages, they provide guidance and the possibility for a world model that is causally aware, which will be essential for Embodied AI.

\paragraph{Causal World Models}
World models, often used in RL, aim to model the transition dynamics which are expected to be intrinsically causal. %
It is a longstanding challenge, where early work has used standard sequence models for this purpose~\cite{schmidhuber1990making}. More recent research has focused on using Variational Autoencoders (VAEs) to map inputs into a latent representation to enable dynamics modelling with high-dimensional observations~\cite{ha2018world, hafner2019learning, gelada2019deepmdp, schrittwieser2020mastering}. Advancements in the Dreamer series \citep{hafner2020mastering, hafner2023mastering} and transformer-based world models \citep{chen2022transdreamer, micheli2022transformers, robine2023transformer} highlight the benefits of discrete latent variables and transformers for latent dynamics. Recent works have also leveraged diffusion models for world models \citep{rigter2023world, anonymous2024diffusion} to improve prediction accuracy at the cost of increased computation. 

The community is increasingly recognizing the important role of causal properties in world models.  Recently, \citet{anonymous2024robust} proved that any agent capable of generalizing under domain shifts must learn a causal model, and there have been many works trying to explicitly incorporate causality. \citet{zhang2020learning,zhang2019learning} use causal core sets \citep{zhang2020invariant} to model latent transition dynamics. \citet{li2020causal, lu2018deconfounding}, on the other hand, considers the confounding effect during the modelling of state transitions. 
Another line of work \citep{annabi2022intrinsically,volodin2020resolving,zhu2022offline,wang2022causal,tomar2021model} explicitly build a SEM among variables in the states, allowing the transition dynamics to only depend on causally-relevant feature. 
Unfortunately, existing causal world models fall short as FWMs due to their inability to generalize across domains and tasks, and to process multi-modal inputs efficiently. Future research on causal-aware FVWM will bridge the gap and enable Embodied AI.

\section{Conclusion}
\label{sec:conclusion}

In this position paper, we advocate for \glspl{fvwm} as key to unlocking the full potential of embodied AI in both general and scalable contexts, highlighting the indispensability of causality in these models. We reiterate that causality as a property is fundamental to accurate action and consequence prediction, and its integration is crucial in the realm of machine learning and foundation models. Looking forward, we envision future research focusing not only on high-modality observations but also on transcending the boundaries of various machine learning sub-fields. This approach aims to optimize data utilization through both physical interactions and a blend of online and offline methodologies. Moreover, we emphasize the significance of establishing empirical evaluation benchmarks to foster real-world impact in a cost-effective and low-risk manner. 

While this paper concentrates on the \glspl{fvwm} in embodied AI, we acknowledge that numerous other research challenges remain unaddressed by this study. These include, but are not limited to, advancements in hardware, human-robot communication and collaboration, and comprehensive socio-economic considerations, which are very important for embodied AI in general.

\vspace{-2mm}
\section*{Impact statement}
This position paper, focusing on world models for Embodied AI, does not propose new technological advancements with direct societal consequences in isolation. However, it is crucial to address the broader social implications of the field. This is a topic of intense debate with a multitude of works predicting its outcome and proposing how to mitigate risks. However, a reoccurring theme in the debate is that is will be necessary to ensure an equitable transition of society to rapidly increasing automation. \citet{willcocks2020robo} argues that such a transition will unlikely result in a net job loss, and will be vital to mitigate changing demographics which are predicted to lead to labour shortages. Using foundation models in Embodied AI shifts the focus of automation from tasks that are dirty, dull, and dangerous (3D) to those that necessitate advanced cognitive and reasoning skills. As a result, the implications of widespread deployment in automating various job types could vary significantly, necessitating careful consideration by society. 

Another important consideration is that Embodied AI should be ethical and fair. For the real-world evaluation of embodied AI, safety and ethical considerations are key, as there is the possibility of inducing physical harm. Moreover, Causal \glspl{fvwm} may reduce biases and lead to fairer algorithms \cite{kusner2017counterfactual,zhang2018fairness}.
However, as suggested by \citet{nilforoshan2022causal}, this may not directly translate to equitable decision making. Instead it will remain important for researchers to develop this area further. 

\section*{Acknowledgment}
The authors wish to express gratitude to Katja Hofmann, Jan Peters, Nan Duan, Anastasia Varava, Xin Tong, Tom Minka,  Jiang Bian, David Sweeney Baining Guo and John Langford, whose discussions have inspired and aided the manuscript.

\bibliography{ref}

\begin{thebibliography}{199}
\providecommand{\natexlab}[1]{#1}
\providecommand{\url}[1]{\texttt{#1}}
\expandafter\ifx\csname urlstyle\endcsname\relax
  \providecommand{\doi}[1]{doi: #1}\else
  \providecommand{\doi}{doi: \begingroup \urlstyle{rm}\Url}\fi

\bibitem[Achiam et~al.(2023)Achiam, Adler, Agarwal, Ahmad, Akkaya, Aleman,
  Almeida, Altenschmidt, Altman, Anadkat, et~al.]{achiam2023gpt}
Achiam, J., Adler, S., Agarwal, S., Ahmad, L., Akkaya, I., Aleman, F.~L.,
  Almeida, D., Altenschmidt, J., Altman, S., Anadkat, S., et~al.
\newblock Gpt-4 technical report.
\newblock \emph{arXiv preprint arXiv:2303.08774}, 2023.

\bibitem[Adams \& Aizawa(2021)Adams and Aizawa]{adams2010causal}
Adams, F. and Aizawa, K.
\newblock {Causal Theories of Mental Content}.
\newblock In Zalta, E.~N. (ed.), \emph{The {Stanford} Encyclopedia of
  Philosophy}. Metaphysics Research Lab, Stanford University, {F}all 2021
  edition, 2021.

\bibitem[Agility(2024)]{digit}
Agility.
\newblock Agility robotics, 2024.
\newblock URL \url{https://agilityrobotics.com/}.

\bibitem[Ahmed et~al.(2020)Ahmed, Tr{\"a}uble, Goyal, Neitz, Bengio,
  Sch{\"o}lkopf, W{\"u}thrich, and Bauer]{ahmed2020causalworld}
Ahmed, O., Tr{\"a}uble, F., Goyal, A., Neitz, A., Bengio, Y., Sch{\"o}lkopf,
  B., W{\"u}thrich, M., and Bauer, S.
\newblock Causalworld: A robotic manipulation benchmark for causal structure
  and transfer learning.
\newblock \emph{arXiv preprint arXiv:2010.04296}, 2020.

\bibitem[Ahn et~al.(2022)Ahn, Brohan, Brown, Chebotar, Cortes, David, Finn, Fu,
  Gopalakrishnan, Hausman, et~al.]{ahn2022can}
Ahn, M., Brohan, A., Brown, N., Chebotar, Y., Cortes, O., David, B., Finn, C.,
  Fu, C., Gopalakrishnan, K., Hausman, K., et~al.
\newblock Do as i can, not as i say: Grounding language in robotic affordances.
\newblock \emph{arXiv preprint arXiv:2204.01691}, 2022.

\bibitem[Ahn et~al.(2024)Ahn, Dwibedi, Finn, Arenas, Gopalakrishnan, Hausman,
  Ichter, Irpan, Joshi, Julian, et~al.]{ahn2024autort}
Ahn, M., Dwibedi, D., Finn, C., Arenas, M.~G., Gopalakrishnan, K., Hausman, K.,
  Ichter, B., Irpan, A., Joshi, N., Julian, R., et~al.
\newblock Autort: Embodied foundation models for large scale orchestration of
  robotic agents.
\newblock \emph{arXiv preprint arXiv:2401.12963}, 2024.

\bibitem[Alonso et~al.(2023)Alonso, Jelley, Kanervisto, and
  Pearce]{anonymous2024diffusion}
Alonso, E., Jelley, A., Kanervisto, A., and Pearce, T.
\newblock Diffusion world models, 2023.
\newblock URL \url{https://openreview.net/forum?id=bAXmvOLtjA}.

\bibitem[Angrist et~al.(1996)Angrist, Imbens, and
  Rubin]{angrist1996identification}
Angrist, J.~D., Imbens, G.~W., and Rubin, D.~B.
\newblock Identification of causal effects using instrumental variables.
\newblock \emph{Journal of the American statistical Association}, 91\penalty0
  (434):\penalty0 444--455, 1996.

\bibitem[Annabi(2022)]{annabi2022intrinsically}
Annabi, L.
\newblock Intrinsically motivated learning of causal world models.
\newblock \emph{arXiv preprint arXiv:2208.04892}, 2022.

\bibitem[Aravind et~al.(2017)Aravind, Raja, and
  P{\'e}rez-Ruiz]{aravind2017task}
Aravind, K.~R., Raja, P., and P{\'e}rez-Ruiz, M.
\newblock Task-based agricultural mobile robots in arable farming: A review.
\newblock \emph{Spanish journal of agricultural research}, 15\penalty0
  (1):\penalty0 e02R01--e02R01, 2017.

\bibitem[Ard{\'o}n et~al.(2020)Ard{\'o}n, Pairet, Lohan, Ramamoorthy, and
  Petrick]{ardon2020affordances}
Ard{\'o}n, P., Pairet, {\`E}., Lohan, K.~S., Ramamoorthy, S., and Petrick, R.
\newblock Affordances in robotic tasks--a survey.
\newblock \emph{arXiv preprint arXiv:2004.07400}, 2020.

\bibitem[Bareinboim(2014)]{bareinboim2014generalizability}
Bareinboim, E.
\newblock \emph{Generalizability in causal inference: Theory and algorithms}.
\newblock PhD thesis, UCLA, 2014.

\bibitem[Bareinboim \& Pearl(2016)Bareinboim and Pearl]{bareinboim2016causal}
Bareinboim, E. and Pearl, J.
\newblock Causal inference and the data-fusion problem.
\newblock \emph{Proceedings of the National Academy of Sciences}, 113\penalty0
  (27):\penalty0 7345--7352, 2016.

\bibitem[Bareinboim et~al.(2022)Bareinboim, Correa, Ibeling, and
  Icard]{bareinboim2022pearl}
Bareinboim, E., Correa, J.~D., Ibeling, D., and Icard, T.
\newblock \emph{On Pearl’s Hierarchy and the Foundations of Causal
  Inference}, pp.\  507–556.
\newblock Association for Computing Machinery, New York, NY, USA, 1 edition,
  2022.
\newblock ISBN 9781450395861.
\newblock URL \url{https://doi.org/10.1145/3501714.3501743}.

\bibitem[Batra et~al.(2020)Batra, Gokaslan, Kembhavi, Maksymets, Mottaghi,
  Savva, Toshev, and Wijmans]{batra2020objectnav}
Batra, D., Gokaslan, A., Kembhavi, A., Maksymets, O., Mottaghi, R., Savva, M.,
  Toshev, A., and Wijmans, E.
\newblock Objectnav revisited: On evaluation of embodied agents navigating to
  objects.
\newblock \emph{arXiv preprint arXiv:2006.13171}, 2020.

\bibitem[Bauer et~al.(2022)Bauer, W{\"u}thrich, Widmaier, Buchholz, Stark,
  Goyal, Steinbrenner, Akpo, Joshi, Berenz, Agrawal, Funk, Urain, Peters,
  Watson, Chen, Srinivasan, Zhang, Zhang, Walter, Madan, Yoneda, Yarats,
  Allshire, Gordon, Bhattacharjee, Srinivasa, Garg, Maeda, Sikchi, Wang, Yao,
  Yang, McCarthy, Sanchez, Wang, Bulens, McGuinness, O'Connor, Stephen, and
  Sch{\"o}lkopf]{rrc2021}
Bauer, S., W{\"u}thrich, M., Widmaier, F., Buchholz, A., Stark, S., Goyal, A.,
  Steinbrenner, T., Akpo, J., Joshi, S., Berenz, V., Agrawal, V., Funk, N.,
  Urain, J., Peters, J., Watson, J., Chen, C., Srinivasan, K., Zhang, J.,
  Zhang, J., Walter, M., Madan, R., Yoneda, T., Yarats, D., Allshire, A.,
  Gordon, E., Bhattacharjee, T., Srinivasa, S., Garg, A., Maeda, T., Sikchi,
  H., Wang, J., Yao, Q., Yang, S., McCarthy, R., Sanchez, F., Wang, Q., Bulens,
  D., McGuinness, K., O'Connor, N., Stephen, R., and Sch{\"o}lkopf, B.
\newblock Real robot challenge: A robotics competition in the cloud.
\newblock In \emph{Proceedings of the NeurIPS 2021 Competitions and
  Demonstrations Track}, volume 176 of \emph{Proceedings of Machine Learning
  Research}, pp.\  190--204. PMLR, 2022.
\newblock URL \url{https://proceedings.mlr.press/v176/bauer22a.html}.

\bibitem[Behley et~al.(2019)Behley, Garbade, Milioto, Quenzel, Behnke,
  Stachniss, and Gall]{behley2019semantickitti}
Behley, J., Garbade, M., Milioto, A., Quenzel, J., Behnke, S., Stachniss, C.,
  and Gall, J.
\newblock Semantickitti: A dataset for semantic scene understanding of lidar
  sequences.
\newblock In \emph{Proceedings of the IEEE/CVF international conference on
  computer vision}, pp.\  9297--9307, 2019.

\bibitem[Berrevoets et~al.(2023)Berrevoets, Kacprzyk, Qian, and van~der
  Schaar]{berrevoets2023causal}
Berrevoets, J., Kacprzyk, K., Qian, Z., and van~der Schaar, M.
\newblock Causal deep learning.
\newblock \emph{arXiv preprint arXiv:2303.02186}, 2023.

\bibitem[Boiko et~al.(2023)Boiko, MacKnight, Kline, and
  Gomes]{boiko2023autonomous}
Boiko, D.~A., MacKnight, R., Kline, B., and Gomes, G.
\newblock Autonomous chemical research with large language models.
\newblock \emph{Nature}, 624\penalty0 (7992):\penalty0 570--578, 2023.

\bibitem[Bommasani et~al.(2021)Bommasani, Hudson, Adeli, Altman, Arora, von
  Arx, Bernstein, Bohg, Bosselut, Brunskill,
  et~al.]{bommasani2021opportunities}
Bommasani, R., Hudson, D.~A., Adeli, E., Altman, R., Arora, S., von Arx, S.,
  Bernstein, M.~S., Bohg, J., Bosselut, A., Brunskill, E., et~al.
\newblock On the opportunities and risks of foundation models.
\newblock \emph{arXiv preprint arXiv:2108.07258}, 2021.

\bibitem[BostonDynamics(2024)]{Spot}
BostonDynamics.
\newblock Spot: The agile mobile robot, 2024.
\newblock URL \url{https://bostondynamics.com/products/spot/}.

\bibitem[Brohan et~al.(2023{\natexlab{a}})Brohan, Brown, Carbajal, Chebotar,
  Chen, Choromanski, Ding, Driess, Dubey, Finn, et~al.]{brohan2023rt}
Brohan, A., Brown, N., Carbajal, J., Chebotar, Y., Chen, X., Choromanski, K.,
  Ding, T., Driess, D., Dubey, A., Finn, C., et~al.
\newblock Rt-2: Vision-language-action models transfer web knowledge to robotic
  control.
\newblock \emph{arXiv preprint arXiv:2307.15818}, 2023{\natexlab{a}}.

\bibitem[Brohan et~al.(2023{\natexlab{b}})Brohan, Chebotar, Finn, Hausman,
  Herzog, Ho, Ibarz, Irpan, Jang, Julian, et~al.]{brohan2023can}
Brohan, A., Chebotar, Y., Finn, C., Hausman, K., Herzog, A., Ho, D., Ibarz, J.,
  Irpan, A., Jang, E., Julian, R., et~al.
\newblock Do as i can, not as i say: Grounding language in robotic affordances.
\newblock In \emph{Conference on Robot Learning}, pp.\  287--318. PMLR,
  2023{\natexlab{b}}.

\bibitem[Bruce et~al.(2024)Bruce, Dennis, Edwards, Parker-Holder, Shi, Hughes,
  Lai, Mavalankar, Steigerwald, Apps, et~al.]{bruce2024genie}
Bruce, J., Dennis, M., Edwards, A., Parker-Holder, J., Shi, Y., Hughes, E.,
  Lai, M., Mavalankar, A., Steigerwald, R., Apps, C., et~al.
\newblock Genie: Generative interactive environments.
\newblock \emph{arXiv preprint arXiv:2402.15391}, 2024.

\bibitem[Bubeck et~al.(2023)Bubeck, Chandrasekaran, Eldan, Gehrke, Horvitz,
  Kamar, Lee, Lee, Li, Lundberg, et~al.]{bubeck2023sparks}
Bubeck, S., Chandrasekaran, V., Eldan, R., Gehrke, J., Horvitz, E., Kamar, E.,
  Lee, P., Lee, Y.~T., Li, Y., Lundberg, S., et~al.
\newblock Sparks of artificial general intelligence: Early experiments with
  gpt-4.
\newblock \emph{arXiv preprint arXiv:2303.12712}, 2023.

\bibitem[Chen et~al.(2022)Chen, Wu, Yoon, and Ahn]{chen2022transdreamer}
Chen, C., Wu, Y.-F., Yoon, J., and Ahn, S.
\newblock Transdreamer: Reinforcement learning with transformer world models.
\newblock \emph{arXiv preprint arXiv:2202.09481}, 2022.

\bibitem[Chen et~al.(2023)Chen, Spiridonova, Yang, Gao, and Li]{chen2023llava}
Chen, W.-G., Spiridonova, I., Yang, J., Gao, J., and Li, C.
\newblock Llava-interactive: An all-in-one demo for image chat, segmentation,
  generation and editing.
\newblock \emph{arXiv preprint arXiv:2311.00571}, 2023.

\bibitem[Chernozhukov et~al.(2018)Chernozhukov, Chetverikov, Demirer, Duflo,
  Hansen, Newey, and Robins]{chernozhukov2018double}
Chernozhukov, V., Chetverikov, D., Demirer, M., Duflo, E., Hansen, C., Newey,
  W., and Robins, J.
\newblock Double/debiased machine learning for treatment and structural
  parameters, 2018.

\bibitem[Cook et~al.(2002)Cook, Campbell, and Shadish]{cook2002experimental}
Cook, T.~D., Campbell, D.~T., and Shadish, W.
\newblock \emph{Experimental and quasi-experimental designs for generalized
  causal inference}, volume 1195.
\newblock Houghton Mifflin Boston, MA, 2002.

\bibitem[Creswell et~al.(2022)Creswell, Shanahan, and
  Higgins]{creswell2022selection}
Creswell, A., Shanahan, M., and Higgins, I.
\newblock Selection-inference: Exploiting large language models for
  interpretable logical reasoning.
\newblock \emph{arXiv preprint arXiv:2205.09712}, 2022.

\bibitem[Cunningham(2021)]{cunningham2021causal}
Cunningham, S.
\newblock Causal inference.
\newblock In \emph{Causal Inference}. Yale University Press, 2021.

\bibitem[Dally et~al.(2021)Dally, Keckler, and Kirk]{dally2021evolution}
Dally, W.~J., Keckler, S.~W., and Kirk, D.~B.
\newblock Evolution of the graphics processing unit (gpu).
\newblock \emph{IEEE Micro}, 41\penalty0 (6):\penalty0 42--51, 2021.

\bibitem[Dennis et~al.(2020)Dennis, Jaques, Vinitsky, Bayen, Russell, Critch,
  and Levine]{dennis2020emergent}
Dennis, M., Jaques, N., Vinitsky, E., Bayen, A., Russell, S., Critch, A., and
  Levine, S.
\newblock Emergent complexity and zero-shot transfer via unsupervised
  environment design.
\newblock \emph{Advances in neural information processing systems},
  33:\penalty0 13049--13061, 2020.

\bibitem[Dorie et~al.(2019)Dorie, Hill, Shalit, Scott, and
  Cervone]{dorie2019automated}
Dorie, V., Hill, J., Shalit, U., Scott, M., and Cervone, D.
\newblock {Automated versus Do-It-Yourself Methods for Causal Inference:
  Lessons Learned from a Data Analysis Competition}.
\newblock \emph{Statistical Science}, 34\penalty0 (1):\penalty0 43 -- 68, 2019.
\newblock \doi{10.1214/18-STS667}.
\newblock URL \url{https://doi.org/10.1214/18-STS667}.

\bibitem[Dzedzickis et~al.(2022)Dzedzickis, Subačiūtė-Žemaitienė,
  Šutinys, Samukaitė-Bubnienė, and
  Bučinskas]{dzedzickis2022industrialrobotics}
Dzedzickis, A., Subačiūtė-Žemaitienė, J., Šutinys, E.,
  Samukaitė-Bubnienė, U., and Bučinskas, V.
\newblock Advanced applications of industrial robotics: New trends and
  possibilities.
\newblock \emph{Applied Sciences}, 12\penalty0 (1), 2022.
\newblock ISSN 2076-3417.
\newblock \doi{10.3390/app12010135}.
\newblock URL \url{https://www.mdpi.com/2076-3417/12/1/135}.

\bibitem[Fernandez-Moral et~al.(2018)Fernandez-Moral, Martins, Wolf, and
  Rives]{fernandez2018new}
Fernandez-Moral, E., Martins, R., Wolf, D., and Rives, P.
\newblock A new metric for evaluating semantic segmentation: leveraging global
  and contour accuracy.
\newblock In \emph{2018 IEEE intelligent vehicles symposium (iv)}, pp.\
  1051--1056. IEEE, 2018.

\bibitem[Firoozi et~al.(2023)Firoozi, Tucker, Tian, Majumdar, Sun, Liu, Zhu,
  Song, Kapoor, Hausman, et~al.]{firoozi2023foundation}
Firoozi, R., Tucker, J., Tian, S., Majumdar, A., Sun, J., Liu, W., Zhu, Y.,
  Song, S., Kapoor, A., Hausman, K., et~al.
\newblock Foundation models in robotics: Applications, challenges, and the
  future.
\newblock \emph{arXiv preprint arXiv:2312.07843}, 2023.

\bibitem[Geffner et~al.(2022)Geffner, Antoran, Foster, Gong, Ma, Kiciman,
  Sharma, Lamb, Kukla, Pawlowski, et~al.]{geffner2022deep}
Geffner, T., Antoran, J., Foster, A., Gong, W., Ma, C., Kiciman, E., Sharma,
  A., Lamb, A., Kukla, M., Pawlowski, N., et~al.
\newblock Deep end-to-end causal inference.
\newblock \emph{arXiv preprint arXiv:2202.02195}, 2022.

\bibitem[Gelada et~al.(2019)Gelada, Kumar, Buckman, Nachum, and
  Bellemare]{gelada2019deepmdp}
Gelada, C., Kumar, S., Buckman, J., Nachum, O., and Bellemare, M.~G.
\newblock Deepmdp: Learning continuous latent space models for representation
  learning.
\newblock In \emph{International Conference on Machine Learning}, pp.\
  2170--2179. PMLR, 2019.

\bibitem[Gibson(1988)]{gibson1988exploratory}
Gibson, E.~J.
\newblock Exploratory behavior in the development of perceiving, acting, and
  the acquiring of knowledge.
\newblock \emph{Annual review of psychology}, 39\penalty0 (1):\penalty0 1--42,
  1988.

\bibitem[Gibson(1978)]{Gibson1978ecological}
Gibson, J.~J.
\newblock The ecological approach to the visual perception of pictures.
\newblock \emph{Leonardo}, 11\penalty0 (3):\penalty0 227--235, 1978.

\bibitem[Glymour et~al.(2019)Glymour, Zhang, and Spirtes]{glymour2019review}
Glymour, C., Zhang, K., and Spirtes, P.
\newblock Review of causal discovery methods based on graphical models.
\newblock \emph{Frontiers in genetics}, 10:\penalty0 524, 2019.

\bibitem[GoldmanSachs(2023)]{GoldmanSachs}
GoldmanSachs.
\newblock Humanoid robots: Sooner than you might think, 2023.
\newblock URL
  \url{https://www.goldmansachs.com/intelligence/pages/humanoid-robots.html}.

\bibitem[Gong et~al.(2019)Gong, Tschiatschek, Nowozin, Turner,
  Hern{\'a}ndez-Lobato, and Zhang]{gong2019icebreaker}
Gong, W., Tschiatschek, S., Nowozin, S., Turner, R.~E., Hern{\'a}ndez-Lobato,
  J.~M., and Zhang, C.
\newblock Icebreaker: Element-wise efficient information acquisition with a
  bayesian deep latent gaussian model.
\newblock \emph{Advances in neural information processing systems}, 32, 2019.

\bibitem[Gong et~al.(2022)Gong, Smith, Wang, Barton, Woodhead, Pawlowski,
  Jennings, and Zhang]{gong2022instructions}
Gong, W., Smith, D., Wang, Z., Barton, C., Woodhead, S., Pawlowski, N.,
  Jennings, J., and Zhang, C.
\newblock Instructions and guide: Causal insights for learning paths in
  education.
\newblock \emph{arXiv preprint arXiv:2208.12610}, 2022.

\bibitem[Gonz{\'a}lez et~al.(2016)Gonz{\'a}lez, Osborne, and
  Lawrence]{gonzalez2016glasses}
Gonz{\'a}lez, J., Osborne, M., and Lawrence, N.
\newblock Glasses: Relieving the myopia of bayesian optimisation.
\newblock In \emph{Artificial Intelligence and Statistics}, pp.\  790--799.
  PMLR, 2016.

\bibitem[Gopnik et~al.(2007)Gopnik, Schulz, and Schulz]{gopnik2007causal}
Gopnik, A., Schulz, L., and Schulz, L.~E.
\newblock \emph{Causal learning: Psychology, philosophy, and computation}.
\newblock Oxford University Press, 2007.

\bibitem[Grau et~al.(2020)Grau, Indri, Bello, and Sauter]{grau2020robots}
Grau, A., Indri, M., Bello, L.~L., and Sauter, T.
\newblock Robots in industry: The past, present, and future of a growing
  collaboration with humans.
\newblock \emph{IEEE Industrial Electronics Magazine}, 15\penalty0
  (1):\penalty0 50--61, 2020.

\bibitem[Green et~al.(2012)Green, Kowalkowski, Paterno, Fischler, Garren, and
  Lu]{green2012art}
Green, C., Kowalkowski, J., Paterno, M., Fischler, M., Garren, L., and Lu, Q.
\newblock The art framework.
\newblock \emph{Journal of Physics: Conference Series}, 396\penalty0
  (2):\penalty0 022020, dec 2012.
\newblock \doi{10.1088/1742-6596/396/2/022020}.
\newblock URL \url{https://dx.doi.org/10.1088/1742-6596/396/2/022020}.

\bibitem[Gunjal et~al.(2023)Gunjal, Yin, and Bas]{gunjal2023detecting}
Gunjal, A., Yin, J., and Bas, E.
\newblock Detecting and preventing hallucinations in large vision language
  models.
\newblock \emph{arXiv preprint arXiv:2308.06394}, 2023.

\bibitem[Gupta et~al.(2022)Gupta, Karkus, Che, Xu, and
  Pavone]{gupta2022foundation}
Gupta, T., Karkus, P., Che, T., Xu, D., and Pavone, M.
\newblock Foundation models for semantic novelty in reinforcement learning.
\newblock \emph{arXiv preprint arXiv:2211.04878}, 2022.

\bibitem[Ha \& Schmidhuber(2018)Ha and Schmidhuber]{ha2018world}
Ha, D. and Schmidhuber, J.
\newblock World models.
\newblock \emph{arXiv preprint arXiv:1803.10122}, 2018.

\bibitem[Hafner et~al.(2019)Hafner, Lillicrap, Fischer, Villegas, Ha, Lee, and
  Davidson]{hafner2019learning}
Hafner, D., Lillicrap, T., Fischer, I., Villegas, R., Ha, D., Lee, H., and
  Davidson, J.
\newblock Learning latent dynamics for planning from pixels.
\newblock In \emph{International conference on machine learning}, pp.\
  2555--2565. PMLR, 2019.

\bibitem[Hafner et~al.(2020)Hafner, Lillicrap, Norouzi, and
  Ba]{hafner2020mastering}
Hafner, D., Lillicrap, T., Norouzi, M., and Ba, J.
\newblock Mastering atari with discrete world models.
\newblock \emph{arXiv preprint arXiv:2010.02193}, 2020.

\bibitem[Hafner et~al.(2023)Hafner, Pasukonis, Ba, and
  Lillicrap]{hafner2023mastering}
Hafner, D., Pasukonis, J., Ba, J., and Lillicrap, T.
\newblock Mastering diverse domains through world models.
\newblock \emph{arXiv preprint arXiv:2301.04104}, 2023.

\bibitem[Hansen et~al.(2022)Hansen, Wang, and Su]{hansen2022temporal}
Hansen, N., Wang, X., and Su, H.
\newblock Temporal difference learning for model predictive control.
\newblock \emph{arXiv preprint arXiv:2203.04955}, 2022.

\bibitem[Herd \& Miles(2019)Herd and Miles]{herd2019detecting}
Herd, B.~C. and Miles, S.
\newblock Detecting causal relationships in simulation models using
  intervention-based counterfactual analysis.
\newblock \emph{ACM Transactions on Intelligent Systems and Technology (TIST)},
  10\penalty0 (5):\penalty0 1--25, 2019.

\bibitem[Hern{\'a}n \& Robins(2010)Hern{\'a}n and Robins]{hernan2010causal}
Hern{\'a}n, M.~A. and Robins, J.~M.
\newblock Causal inference, 2010.

\bibitem[Hoover(2001)]{hoover2001causality}
Hoover, K.~D.
\newblock \emph{Causality in macroeconomics}.
\newblock Cambridge University Press, 2001.

\bibitem[Hu et~al.(2023)Hu, Russell, Yeo, Murez, Fedoseev, Kendall, Shotton,
  and Corrado]{hu2023gaia}
Hu, A., Russell, L., Yeo, H., Murez, Z., Fedoseev, G., Kendall, A., Shotton,
  J., and Corrado, G.
\newblock Gaia-1: A generative world model for autonomous driving.
\newblock \emph{arXiv preprint arXiv:2309.17080}, 2023.

\bibitem[Huang \& Chang(2022)Huang and Chang]{huang2022towards}
Huang, J. and Chang, K. C.-C.
\newblock Towards reasoning in large language models: A survey.
\newblock \emph{arXiv preprint arXiv:2212.10403}, 2022.

\bibitem[Huang et~al.(2023{\natexlab{a}})Huang, Yu, Ma, Zhong, Feng, Wang,
  Chen, Peng, Feng, Qin, et~al.]{huang2023survey}
Huang, L., Yu, W., Ma, W., Zhong, W., Feng, Z., Wang, H., Chen, Q., Peng, W.,
  Feng, X., Qin, B., et~al.
\newblock A survey on hallucination in large language models: Principles,
  taxonomy, challenges, and open questions.
\newblock \emph{arXiv preprint arXiv:2311.05232}, 2023{\natexlab{a}}.

\bibitem[Huang et~al.(2023{\natexlab{b}})Huang, Wang, Zhang, Li, Wu, and
  Fei-Fei]{huang2023voxposer}
Huang, W., Wang, C., Zhang, R., Li, Y., Wu, J., and Fei-Fei, L.
\newblock Voxposer: Composable 3d value maps for robotic manipulation with
  language models.
\newblock \emph{arXiv preprint arXiv:2307.05973}, 2023{\natexlab{b}}.

\bibitem[Imbens(2020)]{imbens2020potential}
Imbens, G.~W.
\newblock Potential outcome and directed acyclic graph approaches to causality:
  Relevance for empirical practice in economics.
\newblock \emph{Journal of Economic Literature}, 58\penalty0 (4):\penalty0
  1129--1179, 2020.

\bibitem[Imbens \& Rubin(2015)Imbens and Rubin]{imbens2015causal}
Imbens, G.~W. and Rubin, D.~B.
\newblock \emph{Causal inference in statistics, social, and biomedical
  sciences}.
\newblock Cambridge University Press, 2015.

\bibitem[Islam et~al.(2022)Islam, Tomar, Lamb, Efroni, Zang, Didolkar, Misra,
  Li, van Seijen, Combes, et~al.]{islam2022agent}
Islam, R., Tomar, M., Lamb, A., Efroni, Y., Zang, H., Didolkar, A., Misra, D.,
  Li, X., van Seijen, H., Combes, R. T.~d., et~al.
\newblock Agent-controller representations: Principled offline rl with rich
  exogenous information.
\newblock \emph{arXiv preprint arXiv:2211.00164}, 2022.

\bibitem[Janner et~al.(2019)Janner, Fu, Zhang, and Levine]{janner2019trust}
Janner, M., Fu, J., Zhang, M., and Levine, S.
\newblock When to trust your model: Model-based policy optimization.
\newblock \emph{Advances in neural information processing systems}, 32, 2019.

\bibitem[Jensen(2019)]{jensen2019comment}
Jensen, D.
\newblock {Comment: Strengthening Empirical Evaluation of Causal Inference
  Methods}.
\newblock \emph{Statistical Science}, 34\penalty0 (1):\penalty0 77 -- 81, 2019.
\newblock \doi{10.1214/18-STS690}.
\newblock URL \url{https://doi.org/10.1214/18-STS690}.

\bibitem[Jin et~al.(2024)Jin, Yu, Zhao, Hua, Meng, Zhang, Du,
  et~al.]{jin2024impact}
Jin, M., Yu, Q., Zhao, H., Hua, W., Meng, Y., Zhang, Y., Du, M., et~al.
\newblock The impact of reasoning step length on large language models.
\newblock \emph{arXiv preprint arXiv:2401.04925}, 2024.

\bibitem[Jin et~al.(2022)Jin, Feder, and Zhang]{jin2022causalnlp}
Jin, Z., Feder, A., and Zhang, K.
\newblock Causalnlp tutorial: An introduction to causality for natural language
  processing.
\newblock In \emph{Proceedings of the 2022 Conference on Empirical Methods in
  Natural Language Processing: Tutorial Abstracts}, pp.\  17--22, 2022.

\bibitem[Jin* et~al.(2023)Jin*, Chen*, Leeb*, Gresele*, Kamal, Lyu, Blin,
  Gonzalez, Kleiman-Weiner, Sachan, and Sch{\"o}lkopf]{Jinetal23}
Jin*, Z., Chen*, Y., Leeb*, F., Gresele*, L., Kamal, O., Lyu, Z., Blin, K.,
  Gonzalez, F., Kleiman-Weiner, M., Sachan, M., and Sch{\"o}lkopf, B.
\newblock {CLadder}: A benchmark to assess causal reasoning capabilities of
  language models.
\newblock In \emph{Advances in Neural Information Processing Systems 36
  (NeurIPS 2023)}, December 2023.

\bibitem[Jin et~al.(2023)Jin, Liu, Lyu, Poff, Sachan, Mihalcea, Diab, and
  Sch{\"o}lkopf]{jin2023can}
Jin, Z., Liu, J., Lyu, Z., Poff, S., Sachan, M., Mihalcea, R., Diab, M., and
  Sch{\"o}lkopf, B.
\newblock Can large language models infer causation from correlation?
\newblock \emph{arXiv preprint arXiv:2306.05836}, 2023.
\newblock ICLR 2024.

\bibitem[Kaddour et~al.(2022)Kaddour, Lynch, Liu, Kusner, and
  Silva]{kaddour2022causal}
Kaddour, J., Lynch, A., Liu, Q., Kusner, M.~J., and Silva, R.
\newblock Causal machine learning: A survey and open problems.
\newblock \emph{arXiv preprint arXiv:2206.15475}, 2022.

\bibitem[Kadian et~al.(2020)Kadian, Truong, Gokaslan, Clegg, Wijmans, Lee,
  Savva, Chernova, and Batra]{kadian2020sim2real}
Kadian, A., Truong, J., Gokaslan, A., Clegg, A., Wijmans, E., Lee, S., Savva,
  M., Chernova, S., and Batra, D.
\newblock Sim2real predictivity: Does evaluation in simulation predict
  real-world performance?
\newblock \emph{IEEE Robotics and Automation Letters}, 5\penalty0 (4):\penalty0
  6670--6677, 2020.

\bibitem[Ke et~al.(2019)Ke, Bilaniuk, Goyal, Bauer, Larochelle, Sch{\"o}lkopf,
  Mozer, Pal, and Bengio]{ke2019learning}
Ke, N.~R., Bilaniuk, O., Goyal, A., Bauer, S., Larochelle, H., Sch{\"o}lkopf,
  B., Mozer, M.~C., Pal, C., and Bengio, Y.
\newblock Learning neural causal models from unknown interventions.
\newblock \emph{arXiv preprint arXiv:1910.01075}, 2019.

\bibitem[Ke et~al.(2022)Ke, Chiappa, Wang, Bornschein, Weber, Goyal, Botvinic,
  Mozer, and Rezende]{ke2022learning}
Ke, N.~R., Chiappa, S., Wang, J., Bornschein, J., Weber, T., Goyal, A.,
  Botvinic, M., Mozer, M., and Rezende, D.~J.
\newblock Learning to induce causal structure.
\newblock \emph{arXiv preprint arXiv:2204.04875}, 2022.

\bibitem[Kearns \& Singh(2002)Kearns and Singh]{kearns2002near}
Kearns, M. and Singh, S.
\newblock Near-optimal reinforcement learning in polynomial time.
\newblock \emph{Machine learning}, 49:\penalty0 209--232, 2002.

\bibitem[Khandelwal et~al.(2022)Khandelwal, Weihs, Mottaghi, and
  Kembhavi]{khandelwal2022simple}
Khandelwal, A., Weihs, L., Mottaghi, R., and Kembhavi, A.
\newblock Simple but effective: Clip embeddings for embodied ai.
\newblock In \emph{Proceedings of the IEEE/CVF Conference on Computer Vision
  and Pattern Recognition}, pp.\  14829--14838, 2022.

\bibitem[Kheddar et~al.(2019)Kheddar, Caron, Gergondet, Comport, Tanguy, Ott,
  Henze, Mesesan, Englsberger, Roa, et~al.]{kheddar2019humanoid}
Kheddar, A., Caron, S., Gergondet, P., Comport, A., Tanguy, A., Ott, C., Henze,
  B., Mesesan, G., Englsberger, J., Roa, M.~A., et~al.
\newblock Humanoid robots in aircraft manufacturing: The airbus use cases.
\newblock \emph{IEEE Robotics \& Automation Magazine}, 26\penalty0
  (4):\penalty0 30--45, 2019.

\bibitem[K{\i}c{\i}man et~al.(2023)K{\i}c{\i}man, Ness, Sharma, and
  Tan]{kiciman2023causal}
K{\i}c{\i}man, E., Ness, R., Sharma, A., and Tan, C.
\newblock Causal reasoning and large language models: Opening a new frontier
  for causality.
\newblock \emph{arXiv preprint arXiv:2305.00050}, 2023.

\bibitem[Killedar et~al.(2021)Killedar, Pokala, and
  Seelamantula]{killedar2021learning}
Killedar, V., Pokala, P.~K., and Seelamantula, C.~S.
\newblock Learning generative prior with latent space sparsity constraints.
\newblock \emph{arXiv preprint arXiv:2105.11956}, 2021.

\bibitem[Kitson et~al.(2021)Kitson, Constantinou, Guo, Liu, and
  Chobtham]{kitson2021survey}
Kitson, N.~K., Constantinou, A.~C., Guo, Z., Liu, Y., and Chobtham, K.
\newblock A survey of bayesian network structure learning.
\newblock \emph{arXiv preprint arXiv:2109.11415}, 2021.

\bibitem[Kjellstr{\"o}m et~al.(2011)Kjellstr{\"o}m, Romero, and
  Kragi{\'c}]{kjellstrom2011visual}
Kjellstr{\"o}m, H., Romero, J., and Kragi{\'c}, D.
\newblock Visual object-action recognition: Inferring object affordances from
  human demonstration.
\newblock \emph{Computer Vision and Image Understanding}, 115\penalty0
  (1):\penalty0 81--90, 2011.

\bibitem[Kolve et~al.(2017)Kolve, Mottaghi, Han, VanderBilt, Weihs, Herrasti,
  Deitke, Ehsani, Gordon, Zhu, et~al.]{kolve2017ai2}
Kolve, E., Mottaghi, R., Han, W., VanderBilt, E., Weihs, L., Herrasti, A.,
  Deitke, M., Ehsani, K., Gordon, D., Zhu, Y., et~al.
\newblock Ai2-thor: An interactive 3d environment for visual ai.
\newblock \emph{arXiv preprint arXiv:1712.05474}, 2017.

\bibitem[Koppula \& Saxena(2013)Koppula and Saxena]{koppula2013anticipating}
Koppula, H.~S. and Saxena, A.
\newblock Anticipating human activities for reactive robotic response.
\newblock In \emph{2013 IEEE/RSJ International Conference on Intelligent Robots
  and Systems}, pp.\  2071--2071, 2013.
\newblock \doi{10.1109/IROS.2013.6696634}.

\bibitem[Koppula et~al.(2013)Koppula, Gupta, and Saxena]{koppula2013learning}
Koppula, H.~S., Gupta, R., and Saxena, A.
\newblock Learning human activities and object affordances from rgb-d videos.
\newblock \emph{The International journal of robotics research}, 32\penalty0
  (8):\penalty0 951--970, 2013.

\bibitem[Koul et~al.(2023)Koul, Sujit, Chen, Evans, Wu, Xu, Chari, Islam,
  Seraj, Efroni, et~al.]{koul2023pclast}
Koul, A., Sujit, S., Chen, S., Evans, B., Wu, L., Xu, B., Chari, R., Islam, R.,
  Seraj, R., Efroni, Y., et~al.
\newblock Pclast: Discovering plannable continuous latent states.
\newblock \emph{arXiv preprint arXiv:2311.03534}, 2023.

\bibitem[Kusner et~al.(2017)Kusner, Loftus, Russell, and
  Silva]{kusner2017counterfactual}
Kusner, M.~J., Loftus, J., Russell, C., and Silva, R.
\newblock Counterfactual fairness.
\newblock \emph{Advances in neural information processing systems}, 30, 2017.

\bibitem[Lamb et~al.(2022)Lamb, Islam, Efroni, Didolkar, Misra, Foster, Molu,
  Chari, Krishnamurthy, and Langford]{lamb2022guaranteed}
Lamb, A., Islam, R., Efroni, Y., Didolkar, A., Misra, D., Foster, D., Molu, L.,
  Chari, R., Krishnamurthy, A., and Langford, J.
\newblock Guaranteed discovery of controllable latent states with multi-step
  inverse models.
\newblock \emph{arXiv preprint arXiv:2207.08229}, 2022.

\bibitem[Lambeta et~al.(2020)Lambeta, Chou, Tian, Yang, Maloon, Most, Stroud,
  Santos, Byagowi, Kammerer, et~al.]{lambeta2020digit}
Lambeta, M., Chou, P.-W., Tian, S., Yang, B., Maloon, B., Most, V.~R., Stroud,
  D., Santos, R., Byagowi, A., Kammerer, G., et~al.
\newblock Digit: A novel design for a low-cost compact high-resolution tactile
  sensor with application to in-hand manipulation.
\newblock \emph{IEEE Robotics and Automation Letters}, 5\penalty0 (3):\penalty0
  3838--3845, 2020.

\bibitem[Lavin et~al.(2021)Lavin, Krakauer, Zenil, Gottschlich, Mattson,
  Brehmer, Anandkumar, Choudry, Rocki, Baydin, et~al.]{lavin2021simulation}
Lavin, A., Krakauer, D., Zenil, H., Gottschlich, J., Mattson, T., Brehmer, J.,
  Anandkumar, A., Choudry, S., Rocki, K., Baydin, A.~G., et~al.
\newblock Simulation intelligence: Towards a new generation of scientific
  methods.
\newblock \emph{arXiv preprint arXiv:2112.03235}, 2021.

\bibitem[Lee et~al.(2022)Lee, Seo, Lee, Abbeel, and Shin]{lee2022offline}
Lee, S., Seo, Y., Lee, K., Abbeel, P., and Shin, J.
\newblock Offline-to-online reinforcement learning via balanced replay and
  pessimistic q-ensemble.
\newblock In \emph{Conference on Robot Learning}, pp.\  1702--1712. PMLR, 2022.

\bibitem[Li et~al.(2023)Li, Gan, Yang, Yang, Li, Wang, and
  Gao]{li2023multimodal}
Li, C., Gan, Z., Yang, Z., Yang, J., Li, L., Wang, L., and Gao, J.
\newblock Multimodal foundation models: From specialists to general-purpose
  assistants.
\newblock \emph{arXiv preprint arXiv:2309.10020}, 1\penalty0 (2):\penalty0 2,
  2023.

\bibitem[Li et~al.(2018)Li, Morgan, and Zaslavsky]{li2018balancing}
Li, F., Morgan, K.~L., and Zaslavsky, A.~M.
\newblock Balancing covariates via propensity score weighting.
\newblock \emph{Journal of the American Statistical Association}, 113\penalty0
  (521):\penalty0 390--400, 2018.

\bibitem[Li \& Han(2024)Li and Han]{li2024enforcing}
Li, H. and Han, T.
\newblock Enforcing sparsity on latent space for robust and explainable
  representations.
\newblock In \emph{Proceedings of the IEEE/CVF Winter Conference on
  Applications of Computer Vision}, pp.\  5282--5291, 2024.

\bibitem[Li et~al.(2020)Li, Yang, Liu, Chen, Chen, and Wang]{li2020causal}
Li, M., Yang, M., Liu, F., Chen, X., Chen, Z., and Wang, J.
\newblock Causal world models by unsupervised deconfounding of physical
  dynamics.
\newblock \emph{arXiv preprint arXiv:2012.14228}, 2020.

\bibitem[Lin et~al.(2023{\natexlab{a}})Lin, Agia, Migimatsu, Pavone, and
  Bohg]{lin2023text2motion}
Lin, K., Agia, C., Migimatsu, T., Pavone, M., and Bohg, J.
\newblock Text2motion: From natural language instructions to feasible plans.
\newblock \emph{arXiv preprint arXiv:2303.12153}, 2023{\natexlab{a}}.

\bibitem[Lin et~al.(2023{\natexlab{b}})Lin, Church, Yang, Li, Lloyd, Zhang, and
  Lepora]{lin2023bi}
Lin, Y., Church, A., Yang, M., Li, H., Lloyd, J., Zhang, D., and Lepora, N.~F.
\newblock Bi-touch: Bimanual tactile manipulation with sim-to-real deep
  reinforcement learning.
\newblock \emph{IEEE Robotics and Automation Letters}, 2023{\natexlab{b}}.

\bibitem[Liu et~al.(2023)Liu, Li, Wu, and Lee]{liu2023visual}
Liu, H., Li, C., Wu, Q., and Lee, Y.~J.
\newblock Visual instruction tuning.
\newblock \emph{arXiv preprint arXiv:2304.08485}, 2023.

\bibitem[Liu et~al.(2024)Liu, Yan, Zaharia, and Abbeel]{liu2024world}
Liu, H., Yan, W., Zaharia, M., and Abbeel, P.
\newblock World model on million-length video and language with ringattention.
\newblock \emph{arXiv preprint arXiv:2402.08268}, 2024.

\bibitem[Lorch et~al.(2022)Lorch, Sussex, Rothfuss, Krause, and
  Sch{\"o}lkopf]{lorch2022amortized}
Lorch, L., Sussex, S., Rothfuss, J., Krause, A., and Sch{\"o}lkopf, B.
\newblock Amortized inference for causal structure learning.
\newblock \emph{Advances in Neural Information Processing Systems},
  35:\penalty0 13104--13118, 2022.

\bibitem[Lu et~al.(2018)Lu, Sch{\"o}lkopf, and
  Hern{\'a}ndez-Lobato]{lu2018deconfounding}
Lu, C., Sch{\"o}lkopf, B., and Hern{\'a}ndez-Lobato, J.~M.
\newblock Deconfounding reinforcement learning in observational settings.
\newblock \emph{arXiv preprint arXiv:1812.10576}, 2018.

\bibitem[Ma \& Zhang(2023)Ma and Zhang]{ma2023high}
Ma, C. and Zhang, C.
\newblock High precision causal model evaluation with conditional
  randomization.
\newblock In \emph{Thirty-seventh Conference on Neural Information Processing
  Systems}, 2023.

\bibitem[Ma et~al.(2019)Ma, Tschiatschek, Palla, Hernandez-Lobato, Nowozin, and
  Zhang]{ma2019eddi}
Ma, C., Tschiatschek, S., Palla, K., Hernandez-Lobato, J.~M., Nowozin, S., and
  Zhang, C.
\newblock Eddi: Efficient dynamic discovery of high-value information with
  partial vae.
\newblock In \emph{International Conference on Machine Learning}, pp.\
  4234--4243. PMLR, 2019.

\bibitem[Ma et~al.(2023)Ma, Liang, Wang, Huang, Bastani, Jayaraman, Zhu, Fan,
  and Anandkumar]{ma2023eureka}
Ma, Y.~J., Liang, W., Wang, G., Huang, D.-A., Bastani, O., Jayaraman, D., Zhu,
  Y., Fan, L., and Anandkumar, A.
\newblock Eureka: Human-level reward design via coding large language models.
\newblock \emph{arXiv preprint arXiv:2310.12931}, 2023.

\bibitem[Makoviychuk et~al.(2021)Makoviychuk, Wawrzyniak, Guo, Lu, Storey,
  Macklin, Hoeller, Rudin, Allshire, Handa, et~al.]{makoviychuk2021isaac}
Makoviychuk, V., Wawrzyniak, L., Guo, Y., Lu, M., Storey, K., Macklin, M.,
  Hoeller, D., Rudin, N., Allshire, A., Handa, A., et~al.
\newblock Isaac gym: High performance gpu-based physics simulation for robot
  learning.
\newblock \emph{arXiv preprint arXiv:2108.10470}, 2021.

\bibitem[Matthieu \& Ga{\"e}l(2023)Matthieu and Ga{\"e}l]{matthieu2023select}
Matthieu, D. and Ga{\"e}l, V.
\newblock How to select predictive models for causal inference?
\newblock \emph{arXiv preprint arXiv:2302.00370}, 2023.

\bibitem[Meyes et~al.(2017)Meyes, Tercan, Roggendorf, Thiele, B{\"u}scher,
  Obdenbusch, Brecher, Jeschke, and Meisen]{meyes2017motion}
Meyes, R., Tercan, H., Roggendorf, S., Thiele, T., B{\"u}scher, C., Obdenbusch,
  M., Brecher, C., Jeschke, S., and Meisen, T.
\newblock Motion planning for industrial robots using reinforcement learning.
\newblock \emph{Procedia CIRP}, 63:\penalty0 107--112, 2017.

\bibitem[Micheli et~al.(2023)Micheli, alonso, and
  Fleuret]{micheli2022transformers}
Micheli, V., alonso, E., and Fleuret, F.
\newblock Transformers are sample efficient world models.
\newblock \emph{International Conference on Learning Representations}, 2023.

\bibitem[Newbury et~al.(2023)Newbury, Gu, Chumbley, Mousavian, Eppner, Leitner,
  Bohg, Morales, Asfour, Kragic, et~al.]{newbury2023deep}
Newbury, R., Gu, M., Chumbley, L., Mousavian, A., Eppner, C., Leitner, J.,
  Bohg, J., Morales, A., Asfour, T., Kragic, D., et~al.
\newblock Deep learning approaches to grasp synthesis: A review.
\newblock \emph{IEEE Transactions on Robotics}, 2023.

\bibitem[Nie \& Wager(2021)Nie and Wager]{nie2021quasi}
Nie, X. and Wager, S.
\newblock Quasi-oracle estimation of heterogeneous treatment effects.
\newblock \emph{Biometrika}, 108\penalty0 (2):\penalty0 299--319, 2021.

\bibitem[Nilforoshan et~al.(2022)Nilforoshan, Gaebler, Shroff, and
  Goel]{nilforoshan2022causal}
Nilforoshan, H., Gaebler, J.~D., Shroff, R., and Goel, S.
\newblock Causal conceptions of fairness and their consequences.
\newblock In \emph{International Conference on Machine Learning}, pp.\
  16848--16887. PMLR, 2022.

\bibitem[Ohneberg et~al.(2023)Ohneberg, St{\"o}bich, Warmbein, Rathgeber,
  Mehler-Klamt, Fischer, and Eberl]{ohneberg2023assistive}
Ohneberg, C., St{\"o}bich, N., Warmbein, A., Rathgeber, I., Mehler-Klamt,
  A.~C., Fischer, U., and Eberl, I.
\newblock Assistive robotic systems in nursing care: a scoping review.
\newblock \emph{BMC nursing}, 22\penalty0 (1):\penalty0 1--15, 2023.

\bibitem[Olsen \& Tyl{\'e}n(2023)Olsen and Tyl{\'e}n]{olsen2023social}
Olsen, K. and Tyl{\'e}n, K.
\newblock On the social nature of abstraction: cognitive implications of
  interaction and diversity.
\newblock \emph{Philosophical Transactions of the Royal Society B},
  378\penalty0 (1870):\penalty0 20210361, 2023.

\bibitem[{Open X-Embodiment Collaboration} et~al.(2023){Open X-Embodiment
  Collaboration}, Padalkar, Pooley, Jain, Bewley, Herzog, Irpan, Khazatsky,
  Rai, Singh, Brohan, Raffin, Wahid, Burgess-Limerick, Kim, Schölkopf, Ichter,
  Lu, Xu, Finn, Xu, Chi, Huang, Chan, Pan, Fu, Devin, Driess, Pathak, Shah,
  Büchler, Kalashnikov, Sadigh, Johns, Ceola, Xia, Stulp, Zhou, Sukhatme,
  Salhotra, Yan, Schiavi, Su, Fang, Shi, Amor, Christensen, Furuta, Walke,
  Fang, Mordatch, Radosavovic, Leal, Liang, Kim, Schneider, Hsu, Bohg, Bingham,
  Wu, Wu, Luo, Gu, Tan, Oh, Malik, Tompson, Yang, Lim, Silvério, Han, Rao,
  Pertsch, Hausman, Go, Gopalakrishnan, Goldberg, Byrne, Oslund, Kawaharazuka,
  Zhang, Majd, Rana, Srinivasan, Chen, Pinto, Tan, Ott, Lee, Tomizuka, Du, Ahn,
  Zhang, Ding, Srirama, Sharma, Kim, Kanazawa, Hansen, Heess, Joshi,
  Suenderhauf, Palo, Shafiullah, Mees, Kroemer, Sanketi, Wohlhart, Xu,
  Sermanet, Sundaresan, Vuong, Rafailov, Tian, Doshi, Martín-Martín,
  Mendonca, Shah, Hoque, Julian, Bustamante, Kirmani, Levine, Moore, Bahl,
  Dass, Song, Xu, Haldar, Adebola, Guist, Nasiriany, Schaal, Welker, Tian,
  Dasari, Belkhale, Osa, Harada, Matsushima, Xiao, Yu, Ding, Davchev, Zhao,
  Armstrong, Darrell, Jain, Vanhoucke, Zhan, Zhou, Burgard, Chen, Wang, Zhu,
  Li, Lu, Chebotar, Zhou, Zhu, Xu, Wang, Bisk, Cho, Lee, Cui, hua Wu, Tang,
  Zhu, Li, Iwasawa, Matsuo, Xu, and Cui]{padalkar2023open}
{Open X-Embodiment Collaboration}, Padalkar, A., Pooley, A., Jain, A., Bewley,
  A., Herzog, A., Irpan, A., Khazatsky, A., Rai, A., Singh, A., Brohan, A.,
  Raffin, A., Wahid, A., Burgess-Limerick, B., Kim, B., Schölkopf, B., Ichter,
  B., Lu, C., Xu, C., Finn, C., Xu, C., Chi, C., Huang, C., Chan, C., Pan, C.,
  Fu, C., Devin, C., Driess, D., Pathak, D., Shah, D., Büchler, D.,
  Kalashnikov, D., Sadigh, D., Johns, E., Ceola, F., Xia, F., Stulp, F., Zhou,
  G., Sukhatme, G.~S., Salhotra, G., Yan, G., Schiavi, G., Su, H., Fang, H.-S.,
  Shi, H., Amor, H.~B., Christensen, H.~I., Furuta, H., Walke, H., Fang, H.,
  Mordatch, I., Radosavovic, I., Leal, I., Liang, J., Kim, J., Schneider, J.,
  Hsu, J., Bohg, J., Bingham, J., Wu, J., Wu, J., Luo, J., Gu, J., Tan, J., Oh,
  J., Malik, J., Tompson, J., Yang, J., Lim, J.~J., Silvério, J., Han, J.,
  Rao, K., Pertsch, K., Hausman, K., Go, K., Gopalakrishnan, K., Goldberg, K.,
  Byrne, K., Oslund, K., Kawaharazuka, K., Zhang, K., Majd, K., Rana, K.,
  Srinivasan, K., Chen, L.~Y., Pinto, L., Tan, L., Ott, L., Lee, L., Tomizuka,
  M., Du, M., Ahn, M., Zhang, M., Ding, M., Srirama, M.~K., Sharma, M., Kim,
  M.~J., Kanazawa, N., Hansen, N., Heess, N., Joshi, N.~J., Suenderhauf, N.,
  Palo, N.~D., Shafiullah, N. M.~M., Mees, O., Kroemer, O., Sanketi, P.~R.,
  Wohlhart, P., Xu, P., Sermanet, P., Sundaresan, P., Vuong, Q., Rafailov, R.,
  Tian, R., Doshi, R., Martín-Martín, R., Mendonca, R., Shah, R., Hoque, R.,
  Julian, R., Bustamante, S., Kirmani, S., Levine, S., Moore, S., Bahl, S.,
  Dass, S., Song, S., Xu, S., Haldar, S., Adebola, S., Guist, S., Nasiriany,
  S., Schaal, S., Welker, S., Tian, S., Dasari, S., Belkhale, S., Osa, T.,
  Harada, T., Matsushima, T., Xiao, T., Yu, T., Ding, T., Davchev, T., Zhao,
  T.~Z., Armstrong, T., Darrell, T., Jain, V., Vanhoucke, V., Zhan, W., Zhou,
  W., Burgard, W., Chen, X., Wang, X., Zhu, X., Li, X., Lu, Y., Chebotar, Y.,
  Zhou, Y., Zhu, Y., Xu, Y., Wang, Y., Bisk, Y., Cho, Y., Lee, Y., Cui, Y., hua
  Wu, Y., Tang, Y., Zhu, Y., Li, Y., Iwasawa, Y., Matsuo, Y., Xu, Z., and Cui,
  Z.~J.
\newblock Open {X-E}mbodiment: Robotic learning datasets and {RT-X} models.
\newblock \url{https://arxiv.org/abs/2310.08864}, 2023.
\newblock ICRA 2024.

\bibitem[Pathak et~al.(2017)Pathak, Agrawal, Efros, and
  Darrell]{pathak2017curiosity}
Pathak, D., Agrawal, P., Efros, A.~A., and Darrell, T.
\newblock Curiosity-driven exploration by self-supervised prediction.
\newblock In \emph{International conference on machine learning}, pp.\
  2778--2787. PMLR, 2017.

\bibitem[Pearl(2009{\natexlab{a}})]{pearl2009causal}
Pearl, J.
\newblock Causal inference in statistics: An overview.
\newblock \emph{Statistics surveys}, 3:\penalty0 96--146, 2009{\natexlab{a}}.

\bibitem[Pearl(2009{\natexlab{b}})]{pearl2009causality}
Pearl, J.
\newblock \emph{Causality}.
\newblock Cambridge university press, 2009{\natexlab{b}}.

\bibitem[Pearl et~al.(2000)]{pearl2000models}
Pearl, J. et~al.
\newblock Causality: models, reasoning and inference.
\newblock \emph{Cambridge, UK: CambridgeUniversityPress}, 19\penalty0
  (2):\penalty0 3, 2000.

\bibitem[Petereit et~al.(2019)Petereit, Beyerer, Asfour, Gentes, Hein,
  Hanebeck, Kirchner, Dillmann, G{\"o}tting, Weiser,
  et~al.]{petereit2019robdekon}
Petereit, J., Beyerer, J., Asfour, T., Gentes, S., Hein, B., Hanebeck, U.~D.,
  Kirchner, F., Dillmann, R., G{\"o}tting, H.~H., Weiser, M., et~al.
\newblock Robdekon: Robotic systems for decontamination in hazardous
  environments.
\newblock In \emph{2019 IEEE International Symposium on Safety, Security, and
  Rescue Robotics (SSRR)}, pp.\  249--255. IEEE, 2019.

\bibitem[Peters et~al.(2017)Peters, Janzing, and
  Sch{\"o}lkopf]{peters2017elements}
Peters, J., Janzing, D., and Sch{\"o}lkopf, B.
\newblock \emph{Elements of causal inference: foundations and learning
  algorithms}.
\newblock The MIT Press, 2017.

\bibitem[Piaget(1965)]{piaget1965stages}
Piaget, J.
\newblock The stages of the intellectual development of the child.
\newblock \emph{Educational psychology in context: Readings for future
  teachers}, 63\penalty0 (4):\penalty0 98--106, 1965.

\bibitem[Puig et~al.(2023)Puig, Undersander, Szot, Cote, Yang, Partsey, Desai,
  Clegg, Hlavac, Min, et~al.]{puig2023habitat}
Puig, X., Undersander, E., Szot, A., Cote, M.~D., Yang, T.-Y., Partsey, R.,
  Desai, R., Clegg, A.~W., Hlavac, M., Min, S.~Y., et~al.
\newblock Habitat 3.0: A co-habitat for humans, avatars and robots.
\newblock \emph{arXiv preprint arXiv:2310.13724}, 2023.

\bibitem[Radford et~al.(2021)Radford, Kim, Hallacy, Ramesh, Goh, Agarwal,
  Sastry, Askell, Mishkin, Clark, et~al.]{radford2021learning}
Radford, A., Kim, J.~W., Hallacy, C., Ramesh, A., Goh, G., Agarwal, S., Sastry,
  G., Askell, A., Mishkin, P., Clark, J., et~al.
\newblock Learning transferable visual models from natural language
  supervision.
\newblock In \emph{International conference on machine learning}, pp.\
  8748--8763. PMLR, 2021.

\bibitem[Rajendran et~al.(2024)Rajendran, Buchholz, Aragam, Sch{\"o}lkopf, and
  Ravikumar]{rajendran2024learning}
Rajendran, G., Buchholz, S., Aragam, B., Sch{\"o}lkopf, B., and Ravikumar, P.
\newblock Learning interpretable concepts: Unifying causal representation
  learning and foundation models.
\newblock \emph{arXiv preprint arXiv:2402.09236}, 2024.

\bibitem[Reed et~al.(2022)Reed, Zolna, Parisotto, Colmenarejo, Novikov,
  Barth-Maron, Gimenez, Sulsky, Kay, Springenberg, et~al.]{reed2022generalist}
Reed, S., Zolna, K., Parisotto, E., Colmenarejo, S.~G., Novikov, A.,
  Barth-Maron, G., Gimenez, M., Sulsky, Y., Kay, J., Springenberg, J.~T.,
  et~al.
\newblock A generalist agent.
\newblock \emph{arXiv preprint arXiv:2205.06175}, 2022.

\bibitem[Ren et~al.(2023)Ren, Ren, Zhang, Buonassisi, and
  Li]{ren2023autonomous}
Ren, Z., Ren, Z., Zhang, Z., Buonassisi, T., and Li, J.
\newblock Autonomous experiments using active learning and ai.
\newblock \emph{Nature Reviews Materials}, 8\penalty0 (9):\penalty0 563--564,
  2023.

\bibitem[Richens \& Everitt(2024)Richens and Everitt]{anonymous2024robust}
Richens, J. and Everitt, T.
\newblock Robust agents learn causal world models.
\newblock In \emph{The Twelfth International Conference on Learning
  Representations}, 2024.
\newblock URL \url{https://openreview.net/forum?id=pOoKI3ouv1}.

\bibitem[Rigter et~al.(2023)Rigter, Yamada, and Posner]{rigter2023world}
Rigter, M., Yamada, J., and Posner, I.
\newblock World models via policy-guided trajectory diffusion.
\newblock \emph{arXiv preprint arXiv:2312.08533}, 2023.

\bibitem[Rigter et~al.(2024)Rigter, Jiang, and Posner]{rigter2024reward}
Rigter, M., Jiang, M., and Posner, I.
\newblock Reward-free curricula for training robust world models.
\newblock \emph{International Conference on Learning Representations}, 2024.

\bibitem[Robine et~al.(2023)Robine, H{\"o}ftmann, Uelwer, and
  Harmeling]{robine2023transformer}
Robine, J., H{\"o}ftmann, M., Uelwer, T., and Harmeling, S.
\newblock Transformer-based world models are happy with 100k interactions.
\newblock \emph{arXiv preprint arXiv:2303.07109}, 2023.

\bibitem[Robins(2000)]{robins2000marginal}
Robins, J.~M.
\newblock Marginal structural models versus structural nested models as tools
  for causal inference.
\newblock In \emph{Statistical models in epidemiology, the environment, and
  clinical trials}, pp.\  95--133. Springer, 2000.

\bibitem[Rolling \& Yang(2014)Rolling and Yang]{rolling2014model}
Rolling, C.~A. and Yang, Y.
\newblock Model selection for estimating treatment effects.
\newblock \emph{Journal of the Royal Statistical Society Series B: Statistical
  Methodology}, 76\penalty0 (4):\penalty0 749--769, 2014.

\bibitem[Rosenbaum \& Rubin(1983)Rosenbaum and Rubin]{rosenbaum1983central}
Rosenbaum, P.~R. and Rubin, D.~B.
\newblock The central role of the propensity score in observational studies for
  causal effects.
\newblock \emph{Biometrika}, 70\penalty0 (1):\penalty0 41--55, 1983.

\bibitem[Rubin(2005)]{rubin2005causal}
Rubin, D.~B.
\newblock Causal inference using potential outcomes: Design, modeling,
  decisions.
\newblock \emph{Journal of the American Statistical Association}, 100\penalty0
  (469):\penalty0 322--331, 2005.

\bibitem[Saito \& Yasui(2020)Saito and Yasui]{saito2020counterfactual}
Saito, Y. and Yasui, S.
\newblock Counterfactual cross-validation: Stable model selection procedure for
  causal inference models.
\newblock In \emph{International Conference on Machine Learning}, pp.\
  8398--8407. PMLR, 2020.

\bibitem[Samvelyan et~al.(2023)Samvelyan, Khan, Dennis, Jiang, Parker-Holder,
  Foerster, Raileanu, and Rockt{\"a}schel]{samvelyan2023maestro}
Samvelyan, M., Khan, A., Dennis, M., Jiang, M., Parker-Holder, J., Foerster,
  J., Raileanu, R., and Rockt{\"a}schel, T.
\newblock Maestro: Open-ended environment design for multi-agent reinforcement
  learning.
\newblock \emph{arXiv preprint arXiv:2303.03376}, 2023.

\bibitem[Sanctuary(2024)]{Phoenix}
Sanctuary.
\newblock Introducing phoenix, 2024.
\newblock URL \url{https://sanctuary.ai/}.

\bibitem[Sanderson(2021)]{stern2021electronic}
Sanderson, K.
\newblock Electronic skin: From flexibility to a sense of touch.
\newblock \emph{Nature}, 591:\penalty0 685, 2021.

\bibitem[Saravanan \& Kouzani(2023)Saravanan and
  Kouzani]{saravanan2023advancements}
Saravanan, K. and Kouzani, A.~Z.
\newblock Advancements in on-device deep neural networks.
\newblock \emph{Information}, 14\penalty0 (8):\penalty0 470, 2023.

\bibitem[Scherrer et~al.(2021)Scherrer, Bilaniuk, Annadani, Goyal, Schwab,
  Sch{\"o}lkopf, Mozer, Bengio, Bauer, and Ke]{scherrer2021learning}
Scherrer, N., Bilaniuk, O., Annadani, Y., Goyal, A., Schwab, P., Sch{\"o}lkopf,
  B., Mozer, M.~C., Bengio, Y., Bauer, S., and Ke, N.~R.
\newblock Learning neural causal models with active interventions.
\newblock \emph{arXiv preprint arXiv:2109.02429}, 2021.

\bibitem[Schmidhuber(1990)]{schmidhuber1990making}
Schmidhuber, J.
\newblock \emph{Making the world differentiable: on using self supervised fully
  recurrent neural networks for dynamic reinforcement learning and planning in
  non-stationary environments}, volume 126.
\newblock Inst. f{\"u}r Informatik, 1990.

\bibitem[Sch\"{o}lkopf(2022)]{Scholkopf22b}
Sch\"{o}lkopf, B.
\newblock \emph{Causality for Machine Learning}, pp.\  765--804.
\newblock Association for Computing Machinery, New York, NY, USA, 1 edition,
  2022.
\newblock ISBN 9781450395861.
\newblock \doi{10.1145/3501714.3501755}.

\bibitem[Sch{\"o}lkopf et~al.(2021)Sch{\"o}lkopf, Locatello, Bauer, Ke,
  Kalchbrenner, Goyal, and Bengio]{scholkopf2021toward}
Sch{\"o}lkopf, B., Locatello, F., Bauer, S., Ke, N.~R., Kalchbrenner, N.,
  Goyal, A., and Bengio, Y.
\newblock Toward causal representation learning.
\newblock \emph{Proceedings of the IEEE}, 109\penalty0 (5):\penalty0 612--634,
  2021.

\bibitem[Schrittwieser et~al.(2020)Schrittwieser, Antonoglou, Hubert, Simonyan,
  Sifre, Schmitt, Guez, Lockhart, Hassabis, Graepel,
  et~al.]{schrittwieser2020mastering}
Schrittwieser, J., Antonoglou, I., Hubert, T., Simonyan, K., Sifre, L.,
  Schmitt, S., Guez, A., Lockhart, E., Hassabis, D., Graepel, T., et~al.
\newblock Mastering atari, go, chess and shogi by planning with a learned
  model.
\newblock \emph{Nature}, 588\penalty0 (7839):\penalty0 604--609, 2020.

\bibitem[Schuler et~al.(2018)Schuler, Baiocchi, Tibshirani, and
  Shah]{schuler2018comparison}
Schuler, A., Baiocchi, M., Tibshirani, R., and Shah, N.
\newblock A comparison of methods for model selection when estimating
  individual treatment effects.
\newblock \emph{arXiv preprint arXiv:1804.05146}, 2018.

\bibitem[Schumann et~al.(2023)Schumann, Zhu, Feng, Fu, Riezler, and
  Wang]{schumann2023velma}
Schumann, R., Zhu, W., Feng, W., Fu, T.-J., Riezler, S., and Wang, W.~Y.
\newblock Velma: Verbalization embodiment of llm agents for vision and language
  navigation in street view.
\newblock \emph{arXiv preprint arXiv:2307.06082}, 2023.

\bibitem[Sekar et~al.(2020)Sekar, Rybkin, Daniilidis, Abbeel, Hafner, and
  Pathak]{sekar2020planning}
Sekar, R., Rybkin, O., Daniilidis, K., Abbeel, P., Hafner, D., and Pathak, D.
\newblock Planning to explore via self-supervised world models.
\newblock In \emph{International Conference on Machine Learning}, pp.\
  8583--8592. PMLR, 2020.

\bibitem[Seo et~al.(2022)Seo, Lee, James, and Abbeel]{seo2022reinforcement}
Seo, Y., Lee, K., James, S.~L., and Abbeel, P.
\newblock Reinforcement learning with action-free pre-training from videos.
\newblock In \emph{International Conference on Machine Learning}, pp.\
  19561--19579. PMLR, 2022.

\bibitem[Shojaie \& Fox(2022)Shojaie and Fox]{shojaie2022granger}
Shojaie, A. and Fox, E.~B.
\newblock Granger causality: A review and recent advances.
\newblock \emph{Annual Review of Statistics and Its Application}, 9:\penalty0
  289--319, 2022.

\bibitem[Shridhar et~al.(2022)Shridhar, Manuelli, and Fox]{shridhar2022cliport}
Shridhar, M., Manuelli, L., and Fox, D.
\newblock Cliport: What and where pathways for robotic manipulation.
\newblock In \emph{Conference on Robot Learning}, pp.\  894--906. PMLR, 2022.

\bibitem[Skreta et~al.(2024)Skreta, Zhou, Yuan, Darvish, Aspuru-Guzik, and
  Garg]{skreta2024replan}
Skreta, M., Zhou, Z., Yuan, J.~L., Darvish, K., Aspuru-Guzik, A., and Garg, A.
\newblock Replan: Robotic replanning with perception and language models.
\newblock \emph{arXiv preprint arXiv:2401.04157}, 2024.

\bibitem[Spirtes \& Zhang(2016)Spirtes and Zhang]{spirtes2016causal}
Spirtes, P. and Zhang, K.
\newblock Causal discovery and inference: concepts and recent methodological
  advances.
\newblock \emph{Applied Informatics}, 3\penalty0 (1):\penalty0 3, Feb 2016.
\newblock ISSN 2196-0089.
\newblock \doi{10.1186/s40535-016-0018-x}.
\newblock URL \url{https://doi.org/10.1186/s40535-016-0018-x}.

\bibitem[Spirtes et~al.(2000)Spirtes, Glymour, and
  Scheines]{spirtes2000causation}
Spirtes, P., Glymour, C.~N., and Scheines, R.
\newblock \emph{Causation, prediction, and search}.
\newblock MIT press, 2000.

\bibitem[Stuart(2010)]{stuart2010matching}
Stuart, E.~A.
\newblock Matching methods for causal inference: A review and a look forward.
\newblock \emph{Statistical science: a review journal of the Institute of
  Mathematical Statistics}, 25\penalty0 (1):\penalty0 1, 2010.

\bibitem[Sun et~al.(2022)Sun, Kuchenbecker, and Martius]{sun2022soft}
Sun, H., Kuchenbecker, K.~J., and Martius, G.
\newblock A soft thumb-sized vision-based sensor with accurate all-round force
  perception.
\newblock \emph{Nature Machine Intelligence}, 4\penalty0 (2):\penalty0
  135--145, 2022.

\bibitem[Sun et~al.(2023)Sun, Zheng, Xie, Liu, Chu, Qiu, Xu, Ding, Li, Geng,
  et~al.]{sun2023survey}
Sun, J., Zheng, C., Xie, E., Liu, Z., Chu, R., Qiu, J., Xu, J., Ding, M., Li,
  H., Geng, M., et~al.
\newblock A survey of reasoning with foundation models.
\newblock \emph{arXiv preprint arXiv:2312.11562}, 2023.

\bibitem[Sutton(1991)]{sutton1991dyna}
Sutton, R.~S.
\newblock Dyna, an integrated architecture for learning, planning, and
  reacting.
\newblock \emph{ACM Sigart Bulletin}, 2\penalty0 (4):\penalty0 160--163, 1991.

\bibitem[Tam et~al.(2022)Tam, Rabinowitz, Lampinen, Roy, Chan, Strouse, Wang,
  Banino, and Hill]{tam2022semantic}
Tam, A., Rabinowitz, N., Lampinen, A., Roy, N.~A., Chan, S., Strouse, D., Wang,
  J., Banino, A., and Hill, F.
\newblock Semantic exploration from language abstractions and pretrained
  representations.
\newblock \emph{Advances in neural information processing systems},
  35:\penalty0 25377--25389, 2022.

\bibitem[Tilley(2017)]{tilley2017automation}
Tilley, J.
\newblock Automation, robotics, and the factory of the future.
\newblock \emph{McKinsey \& Company}, 67\penalty0 (1):\penalty0 67--72, 2017.

\bibitem[Tomar et~al.(2021)Tomar, Zhang, Calandra, Taylor, and
  Pineau]{tomar2021model}
Tomar, M., Zhang, A., Calandra, R., Taylor, M.~E., and Pineau, J.
\newblock Model-invariant state abstractions for model-based reinforcement
  learning.
\newblock \emph{arXiv preprint arXiv:2102.09850}, 2021.

\bibitem[Toth et~al.(2022)Toth, Lorch, Knoll, Krause, Pernkopf, Peharz, and von
  K{\"u}gelgen]{toth2022active}
Toth, C., Lorch, L., Knoll, C., Krause, A., Pernkopf, F., Peharz, R., and von
  K{\"u}gelgen, J.
\newblock Active bayesian causal inference.
\newblock \emph{arXiv preprint arXiv:2206.02063}, 2022.

\bibitem[Trujillo~Herrera \& Labram(2020)Trujillo~Herrera and
  Labram]{trujillo2020perovskite}
Trujillo~Herrera, C. and Labram, J.~G.
\newblock A perovskite retinomorphic sensor.
\newblock \emph{Applied Physics Letters}, 117\penalty0 (23), 2020.

\bibitem[Tu et~al.(2023)Tu, Ma, and Zhang]{tu2023causal}
Tu, R., Ma, C., and Zhang, C.
\newblock Causal-discovery performance of chatgpt in the context of neuropathic
  pain diagnosis.
\newblock \emph{arXiv preprint arXiv:2301.13819}, 2023.

\bibitem[Unitree(2024)]{Go2}
Unitree.
\newblock Unitree go2, 2024.
\newblock URL \url{https://m.unitree.com/go2/}.

\bibitem[Vansteelandt et~al.(2012)Vansteelandt, Bekaert, and
  Claeskens]{vansteelandt2012model}
Vansteelandt, S., Bekaert, M., and Claeskens, G.
\newblock On model selection and model misspecification in causal inference.
\newblock \emph{Statistical methods in medical research}, 21\penalty0
  (1):\penalty0 7--30, 2012.

\bibitem[Volodin et~al.(2020)Volodin, Wichers, and Nixon]{volodin2020resolving}
Volodin, S., Wichers, N., and Nixon, J.
\newblock Resolving spurious correlations in causal models of environments via
  interventions.
\newblock \emph{arXiv preprint arXiv:2002.05217}, 2020.

\bibitem[Wager \& Athey(2018)Wager and Athey]{wager2018estimation}
Wager, S. and Athey, S.
\newblock Estimation and inference of heterogeneous treatment effects using
  random forests.
\newblock \emph{Journal of the American Statistical Association}, 113\penalty0
  (523):\penalty0 1228--1242, 2018.

\bibitem[Wang et~al.(2023{\natexlab{a}})Wang, Xu, Lan, Hu, Lan, Lee, and
  Lim]{wang2023plan}
Wang, L., Xu, W., Lan, Y., Hu, Z., Lan, Y., Lee, R. K.-W., and Lim, E.-P.
\newblock Plan-and-solve prompting: Improving zero-shot chain-of-thought
  reasoning by large language models.
\newblock \emph{arXiv preprint arXiv:2305.04091}, 2023{\natexlab{a}}.

\bibitem[Wang \& Jordan(2021)Wang and Jordan]{wang2021desiderata}
Wang, Y. and Jordan, M.~I.
\newblock Desiderata for representation learning: A causal perspective.
\newblock \emph{arXiv preprint arXiv:2109.03795}, 2021.

\bibitem[Wang et~al.(2022)Wang, Xiao, Xu, Zhu, and Stone]{wang2022causal}
Wang, Z., Xiao, X., Xu, Z., Zhu, Y., and Stone, P.
\newblock Causal dynamics learning for task-independent state abstraction.
\newblock \emph{arXiv preprint arXiv:2206.13452}, 2022.

\bibitem[Wang et~al.(2023{\natexlab{b}})Wang, Prabha, Huang, Wu, and
  Rajagopal]{wang2023skyscript}
Wang, Z., Prabha, R., Huang, T., Wu, J., and Rajagopal, R.
\newblock Skyscript: A large and semantically diverse vision-language dataset
  for remote sensing.
\newblock \emph{arXiv preprint arXiv:2312.12856}, 2023{\natexlab{b}}.

\bibitem[Ward-Cherrier et~al.(2018)Ward-Cherrier, Pestell, Cramphorn, Winstone,
  Giannaccini, Rossiter, and Lepora]{ward2018tactip}
Ward-Cherrier, B., Pestell, N., Cramphorn, L., Winstone, B., Giannaccini,
  M.~E., Rossiter, J., and Lepora, N.~F.
\newblock The tactip family: Soft optical tactile sensors with 3d-printed
  biomimetic morphologies.
\newblock \emph{Soft robotics}, 5\penalty0 (2):\penalty0 216--227, 2018.

\bibitem[Webb et~al.(2023)Webb, Holyoak, and Lu]{webb2023emergent}
Webb, T., Holyoak, K.~J., and Lu, H.
\newblock Emergent analogical reasoning in large language models.
\newblock \emph{Nature Human Behaviour}, 7\penalty0 (9):\penalty0 1526--1541,
  2023.

\bibitem[Wei et~al.(2022)Wei, Wang, Schuurmans, Bosma, Xia, Chi, Le, Zhou,
  et~al.]{wei2022chain}
Wei, J., Wang, X., Schuurmans, D., Bosma, M., Xia, F., Chi, E., Le, Q.~V.,
  Zhou, D., et~al.
\newblock Chain-of-thought prompting elicits reasoning in large language
  models.
\newblock \emph{Advances in Neural Information Processing Systems},
  35:\penalty0 24824--24837, 2022.

\bibitem[Willcocks(2020)]{willcocks2020robo}
Willcocks, L.
\newblock Robo-apocalypse cancelled? reframing the automation and future of
  work debate.
\newblock \emph{Journal of Information Technology}, 35\penalty0 (4):\penalty0
  286--302, 2020.

\bibitem[Willig et~al.(2022)Willig, Ze{\v{c}}evi{\'c}, Dhami, and
  Kersting]{willig2022can}
Willig, M., Ze{\v{c}}evi{\'c}, M., Dhami, D.~S., and Kersting, K.
\newblock Can foundation models talk causality?
\newblock \emph{arXiv preprint arXiv:2206.10591}, 2022.

\bibitem[Xiao et~al.(2023)Xiao, Liu, Wang, Zhou, Qi, Cheng, He, and
  Jiang]{xiao2023robot}
Xiao, X., Liu, J., Wang, Z., Zhou, Y., Qi, Y., Cheng, Q., He, B., and Jiang, S.
\newblock Robot learning in the era of foundation models: A survey.
\newblock \emph{arXiv preprint arXiv:2311.14379}, 2023.

\bibitem[Yang et~al.(2023)Yang, Tan, Jin, Liu, Fu, Song, and
  Wang]{yang2023pave}
Yang, J., Tan, W., Jin, C., Liu, B., Fu, J., Song, R., and Wang, L.
\newblock Pave the way to grasp anything: Transferring foundation models for
  universal pick-place robots.
\newblock \emph{arXiv preprint arXiv:2306.05716}, 2023.

\bibitem[Yin et~al.(2020)Yin, Li, Pan, Zhang, and
  Tschiatschek]{yin2020reinforcement}
Yin, H., Li, Y., Pan, S.~J., Zhang, C., and Tschiatschek, S.
\newblock Reinforcement learning with efficient active feature acquisition.
\newblock \emph{arXiv preprint arXiv:2011.00825}, 2020.

\bibitem[Yin et~al.(2021)Yin, Varava, and Kragic]{yin2021modeling}
Yin, H., Varava, A., and Kragic, D.
\newblock Modeling, learning, perception, and control methods for deformable
  object manipulation.
\newblock \emph{Science Robotics}, 6\penalty0 (54):\penalty0 eabd8803, 2021.

\bibitem[Yin et~al.(2023)Yin, Wu, Yang, Wang, Wang, Ni, Yang, Li, Liu, Yang,
  et~al.]{yin2023nuwa}
Yin, S., Wu, C., Yang, H., Wang, J., Wang, X., Ni, M., Yang, Z., Li, L., Liu,
  S., Yang, F., et~al.
\newblock Nuwa-xl: Diffusion over diffusion for extremely long video
  generation.
\newblock \emph{arXiv preprint arXiv:2303.12346}, 2023.

\bibitem[Yuan et~al.(2021)Yuan, Chen, Chen, Codella, Dai, Gao, Hu, Huang, Li,
  Li, et~al.]{yuan2021florence}
Yuan, L., Chen, D., Chen, Y.-L., Codella, N., Dai, X., Gao, J., Hu, H., Huang,
  X., Li, B., Li, C., et~al.
\newblock Florence: A new foundation model for computer vision.
\newblock \emph{arXiv preprint arXiv:2111.11432}, 2021.

\bibitem[Zahid \& Pokorny(2023)Zahid and Pokorny]{zahid2023cloudgripper}
Zahid, M. and Pokorny, F.~T.
\newblock Cloudgripper: An open source cloud robotics testbed for robotic
  manipulation research, benchmarking and data collection at scale.
\newblock \emph{arXiv preprint arXiv:2309.12786}, 2023.

\bibitem[Zanga et~al.(2022)Zanga, Ozkirimli, and Stella]{zanga2022survey}
Zanga, A., Ozkirimli, E., and Stella, F.
\newblock A survey on causal discovery: Theory and practice.
\newblock \emph{International Journal of Approximate Reasoning}, 151:\penalty0
  101--129, 2022.

\bibitem[Ze{\v{c}}evi{\'c} et~al.(2023)Ze{\v{c}}evi{\'c}, Willig, Dhami, and
  Kersting]{zevcevic2023causal}
Ze{\v{c}}evi{\'c}, M., Willig, M., Dhami, D.~S., and Kersting, K.
\newblock Causal parrots: Large language models may talk causality but are not
  causal.
\newblock \emph{arXiv preprint arXiv:2308.13067}, 2023.

\bibitem[Zeng et~al.(2023)Zeng, Cai, Sun, Huang, and Hao]{zeng2023survey}
Zeng, Y., Cai, R., Sun, F., Huang, L., and Hao, Z.
\newblock A survey on causal reinforcement learning, 2023.

\bibitem[Zhang et~al.(2019)Zhang, Lipton, Pineda, Azizzadenesheli, Anandkumar,
  Itti, Pineau, and Furlanello]{zhang2019learning}
Zhang, A., Lipton, Z.~C., Pineda, L., Azizzadenesheli, K., Anandkumar, A.,
  Itti, L., Pineau, J., and Furlanello, T.
\newblock Learning causal state representations of partially observable
  environments.
\newblock \emph{arXiv preprint arXiv:1906.10437}, 2019.

\bibitem[Zhang et~al.(2020{\natexlab{a}})Zhang, Lyle, Sodhani, Filos,
  Kwiatkowska, Pineau, Gal, and Precup]{zhang2020invariant}
Zhang, A., Lyle, C., Sodhani, S., Filos, A., Kwiatkowska, M., Pineau, J., Gal,
  Y., and Precup, D.
\newblock Invariant causal prediction for block mdps.
\newblock In \emph{International Conference on Machine Learning}, pp.\
  11214--11224. PMLR, 2020{\natexlab{a}}.

\bibitem[Zhang et~al.(2020{\natexlab{b}})Zhang, McAllister, Calandra, Gal, and
  Levine]{zhang2020learning}
Zhang, A., McAllister, R., Calandra, R., Gal, Y., and Levine, S.
\newblock Learning invariant representations for reinforcement learning without
  reconstruction.
\newblock \emph{arXiv preprint arXiv:2006.10742}, 2020{\natexlab{b}}.

\bibitem[Zhang et~al.(2020{\natexlab{c}})Zhang, Zhang, and Li]{zhang2020causal}
Zhang, C., Zhang, K., and Li, Y.
\newblock A causal view on robustness of neural networks.
\newblock \emph{Advances in Neural Information Processing Systems},
  33:\penalty0 289--301, 2020{\natexlab{c}}.

\bibitem[Zhang et~al.(2023{\natexlab{a}})Zhang, Bauer, Bennett, Gao, Gong,
  Hilmkil, Jennings, Ma, Minka, Pawlowski, et~al.]{zhang2023understanding}
Zhang, C., Bauer, S., Bennett, P., Gao, J., Gong, W., Hilmkil, A., Jennings,
  J., Ma, C., Minka, T., Pawlowski, N., et~al.
\newblock Understanding causality with large language models: Feasibility and
  opportunities.
\newblock \emph{arXiv preprint arXiv:2304.05524}, 2023{\natexlab{a}}.

\bibitem[Zhang et~al.(2023{\natexlab{b}})Zhang, Janzing, van~der Schaar,
  Locatello, Spirtes, Zhang, {Sch\"olkopf}, and Uhler]{zhang2023causality}
Zhang, C., Janzing, D., van~der Schaar, M., Locatello, F., Spirtes, P., Zhang,
  K., {Sch\"olkopf}, B., and Uhler, C.
\newblock Causality in the time of {LLMs}: Round table discussion results of
  {CLeaR} 2023, 2023{\natexlab{b}}.
\newblock URL
  \url{https://www.cclear.cc/2023/CLeaR23_roundtable_discussion.pdf}.

\bibitem[Zhang \& Bareinboim(2018)Zhang and Bareinboim]{zhang2018fairness}
Zhang, J. and Bareinboim, E.
\newblock Fairness in decision-making—the causal explanation formula.
\newblock In \emph{Proceedings of the AAAI Conference on Artificial
  Intelligence}, volume~32, 2018.

\bibitem[Zhang et~al.(2023{\natexlab{c}})Zhang, Jennings, Zhang, and
  Ma]{zhang2023towards}
Zhang, J., Jennings, J., Zhang, C., and Ma, C.
\newblock Towards causal foundation model: on duality between causal inference
  and attention.
\newblock \emph{arXiv preprint arXiv:2310.00809}, 2023{\natexlab{c}}.

\bibitem[Zhang et~al.(2022)Zhang, Zhang, Li, and Smola]{zhang2022automatic}
Zhang, Z., Zhang, A., Li, M., and Smola, A.
\newblock Automatic chain of thought prompting in large language models.
\newblock \emph{arXiv preprint arXiv:2210.03493}, 2022.

\bibitem[Zhou et~al.(2023{\natexlab{a}})Zhou, Li, Li, Yu, Liu, Wang, Zhang, Ji,
  Yan, He, et~al.]{zhou2023comprehensive}
Zhou, C., Li, Q., Li, C., Yu, J., Liu, Y., Wang, G., Zhang, K., Ji, C., Yan,
  Q., He, L., et~al.
\newblock A comprehensive survey on pretrained foundation models: A history
  from bert to chatgpt.
\newblock \emph{arXiv preprint arXiv:2302.09419}, 2023{\natexlab{a}}.

\bibitem[Zhou et~al.(2023{\natexlab{b}})Zhou, Yao, Xu, Wang, Zhu, and
  Zhang]{zhou2023opportunity}
Zhou, G., Yao, L., Xu, X., Wang, C., Zhu, L., and Zhang, K.
\newblock On the opportunity of causal deep generative models: A survey and
  future directions.
\newblock \emph{arXiv preprint arXiv:2301.12351}, 2023{\natexlab{b}}.

\bibitem[Zhu et~al.(2022)Zhu, Chen, Tian, Zhang, and Yu]{zhu2022offline}
Zhu, Z.-M., Chen, X.-H., Tian, H.-L., Zhang, K., and Yu, Y.
\newblock Offline reinforcement learning with causal structured world models.
\newblock \emph{arXiv preprint arXiv:2206.01474}, 2022.

\end{thebibliography}
\bibliographystyle{icml2024}

\end{document}